\documentclass[]{ailab}
\usepackage{amsmath,amssymb}
\usepackage{array,tabularx,enumitem}
\usepackage{colortbl}
\usepackage{algorithm}
\usepackage{algpseudocode}
\usepackage{tikz}
\usetikzlibrary{arrows.meta,positioning,fit,calc}
\newcolumntype{Y}{>{\raggedright\arraybackslash}X}
\newcommand{\method}{\textsc{RiseHarness}}
\newcommand{\code}[1]{\texttt{\detokenize{#1}}}
\newcommand{\algop}[1]{\mathop{\text{\normalfont\fontfamily{pcr}\bfseries\selectfont #1}}\nolimits}
\newcommand{\ind}{\mathbb{I}}
\newcommand{\pending}{\textit{Pending}}
\definecolor{gapcoral}{HTML}{F18479}
\newcommand{\resultgap}[2]{\mbox{#1\,\textcolor{gapcoral}{\raisebox{0.55ex}{\fontfamily{\sfdefault}\fontseries{bx}\fontsize{6.5}{7}\selectfont #2}}}}
\setlist[itemize]{leftmargin=*,itemsep=2pt,topsep=4pt}
\setlist[enumerate]{leftmargin=*,itemsep=2pt,topsep=4pt}
\hypersetup{pdftitle={Harness Self-Evolution for Long-Horizon Terminal Agents},
  pdfauthor={Ruhan Wang, Yucheng Shi, Zongxia Li, Junyao Yang, Zhongzhi Li, Yue Yu, Haitao Mi, Dongruo Zhou, Leoweiliang},pdfsubject={Working research draft}}

\title{Harness Self-Evolution for Long-Horizon Terminal Agents}
\author[1,2,\ast]{Ruhan Wang}
\author[1,\dagger]{Yucheng Shi}
\author[1,3]{Zongxia Li}
\author[1,5]{Junyao Yang}
\author[1,4]{Zhongzhi Li}
\author[2]{Yue Yu}
\author[1]{Haitao Mi}
\author[2]{Dongruo Zhou}
\author[1]{Leoweiliang}
\affiliation[1]{Tencent HY LLM Frontier}
\affiliation[2]{Indiana University}
\affiliation[3]{University of Maryland, College Park}
\affiliation[4]{University of Georgia}
\affiliation[5]{National University of Singapore}
\contribution[\ast]{This work was conducted during an internship at Tencent.}
\contribution[\dagger]{Internship mentor.}
\date{October 19, 2026}
\abstract{
Long-horizon terminal agents may complete a task in one run yet fail to do so
consistently across repeated runs. Stochastic generation can introduce local
errors that propagate through subsequent actions and context, creating a gap
between task-solving ability and reliable execution. The harness controls
command execution, feedback, context management, and completion decisions,
providing a direct way to address this gap. The effectiveness of these
mechanisms depends on the model's behavior. We present \method{}, a framework for iteratively
improving agent harnesses while keeping model weights fixed. \method{} focuses
on \emph{flipping tasks}, which have both successful and failed runs under the
same configuration. Successful runs establish that the current system can
complete the task and provide references for analyzing failures. The analysis
combines agent trajectories with internal harness traces collected by a
runtime monitor, examining both the model's actions and how the harness
executes them, presents feedback, and controls the interaction. A Harness
Handbook connects this evidence to specific harness behaviors and their source
code, helping identify recurring problems in mechanisms shared across tasks.
To address these problems, \method{} searches other harness frameworks for
relevant implementations and uses them as design references, taking into
account the current execution flow, interaction protocol, available signals,
and model behavior. Candidate changes undergo
behavioral checks and repeated task evaluation; versions with measured gains
become the basis for the next iteration. This loop progressively integrates
suitable mechanisms from different frameworks into a single harness and uses
new execution evidence to guide further changes. We evaluate task success and
reliability across models, use ablations to examine the contributions of
harness execution evidence and external research, and test
transfer to unseen benchmarks with the same model. The framework treats a
specific model as the unit of harness adaptation, studying how
changes to shared interaction mechanisms can make existing capabilities more
reliable across tasks.
}

\begin{document}
\thispagestyle{firstheader}
\maketitle

\section{Introduction}
\label{sec:introduction}

A model that solves a long-horizon terminal task once may fail to do so
consistently. Across many interactions, an incorrect command or a missed
observation can change the environment and the context available for later
decisions. Stochastic generation can therefore lead to different outcomes even
when the task and model remain unchanged. Long-horizon terminal benchmarks
examine workflows requiring hundreds of interactions and sustained progress
\citep{li2026longhorizon,li2026lhtb2}. This distinction between reaching a
successful outcome and succeeding reliably across repeated runs is central to
agent evaluation \citep{yao2024taubench,rabanser2026reliability}. We study how to narrow this gap by
improving the harness around a fixed model.

Training-based approaches improve terminal agents through reinforcement
learning and verified task synthesis \citep{yang2026t1,li2026recursivesynthesis}.
The harness controls command execution, observation feedback, context
management, and completion decisions. Interface design can affect agent
performance without updating model weights \citep{yang2024sweagent}, and
the harness itself has become a target for automatic improvement. Execution
traces and evaluation feedback can guide changes to harness code and support
the evaluation of their effects
\citep{lee2026metaharness,lin2026ahe,chen2026harnessfix}. Failure-informed
runtime interventions have also been studied as a way to improve repeated
success without changing model weights \citep{agarwal2026fire}.

Our focus is how to connect differences between successful and failed runs to
actual internal harness behavior, identify shared mechanisms to improve, and
use implementations from other frameworks as references for changes suited to
the current harness and model. A difference between two trajectories does not
by itself establish that a harness behavior is missing: the harness may
already provide the relevant feedback, or the information needed for a
proposed intervention may not be available. Likewise, an effective design in
another framework may depend on different interfaces or execution conditions.
We introduce \method{}, which uses execution evidence to check these
conditions and to design and validate changes. With the evaluated model held
fixed, its evolution loop combines comparisons of successful and failed runs,
internal execution analysis, and research into existing implementations.

We begin with \emph{flipping tasks}, which have both successful and failed
runs under the same model and harness configuration. Successful runs show that
the current system can solve the task and provide references for analyzing
failures. Same-task trajectory contrast and cross-task aggregation
help identify recurring behavioral problems \citep{wu2026modularrsi}.
In \method{}, the
contrast first identifies a behavioral need; the analysis then checks whether
the current harness can support that behavior and whether it already performs
it. A runtime monitor records internal execution and aligns these harness
traces with agent trajectories, exposing how proposed actions are dispatched
and how feedback reaches the model. We use the Harness Handbook
\citep{wang2026handbook} to connect these records to behavioral responsibilities
and implementing code. This joint analysis helps distinguish an agent's
unsuccessful response from a missing or ineffective harness mechanism and
groups supported findings into targets for shared changes.

External documentation can inform the implementation of agent capabilities
\citep{cheng2026mem2evolve}. \method{} searches other frameworks, source code,
and technical documentation for implementations relevant to the behavioral
problems identified through execution analysis. For completion decisions, for
example, it can examine how other frameworks track unfinished commands or
verify results. The system studies the signals and interfaces these designs
require, then considers the current harness's execution flow, available
information, and model behavior to decide which ideas are applicable and how
to modify existing mechanisms. The goal is to draw on useful designs from
different frameworks to develop a harness whose mechanisms work together for
the current model.

Behavioral checks assess whether a candidate implements the intended change,
while repeated task evaluations measure its effect on performance. Accepted
changes provide the basis for subsequent rounds; unsuccessful candidates
supply execution evidence for revising the implementation or proposed
mechanism. Model weights remain fixed throughout.

\method{} evolves the harness for a specific model based on its execution
behavior. Models may require different forms of interaction support
\citep{luo2026harnessbank,wu2026harnessdev}; we therefore conduct evolution
separately for each model and examine the resulting behavioral changes.
These changes affect shared mechanisms for command execution, observation
feedback, and completion. Their benefits beyond the evolution tasks require
separate validation. Repeated evaluations and ablations assess the effects on task
success and execution reliability, as well as the contributions of dual-trace
evidence and research into external implementations. We then hold both the model and
the evolved harness fixed and evaluate on benchmarks not used for evolution
to test whether these behavioral improvements support a broader range of tasks.

Our main contributions are:
\begin{itemize}
  \item \textbf{An evidence-driven harness evolution framework.}
  We introduce \method{}, which uses successful runs as references for failed
  runs and iteratively retains harness changes that improve measured
  performance without updating model weights.
  \item \textbf{Behavioral diagnosis from paired runs and internal execution.}
  We combine agent and harness traces with the Harness Handbook to check
  proposed behavioral needs against actual execution and locate shared
  mechanisms for modification.
  \item \textbf{Harness modification informed by external implementations.}
  We retrieve relevant implementations from other frameworks, examine their
  requirements, and use the current harness's structure, runtime signals, and
  model behavior to design specific changes.
  \item \textbf{Reliability and generalization.}
  Repeated runs and ablations assess harness evidence and external research;
  we compare changes across models and test cross-benchmark transfer with
  the model held fixed.
\end{itemize}



\begin{figure}[t]
\centering
\includegraphics[width=\linewidth]{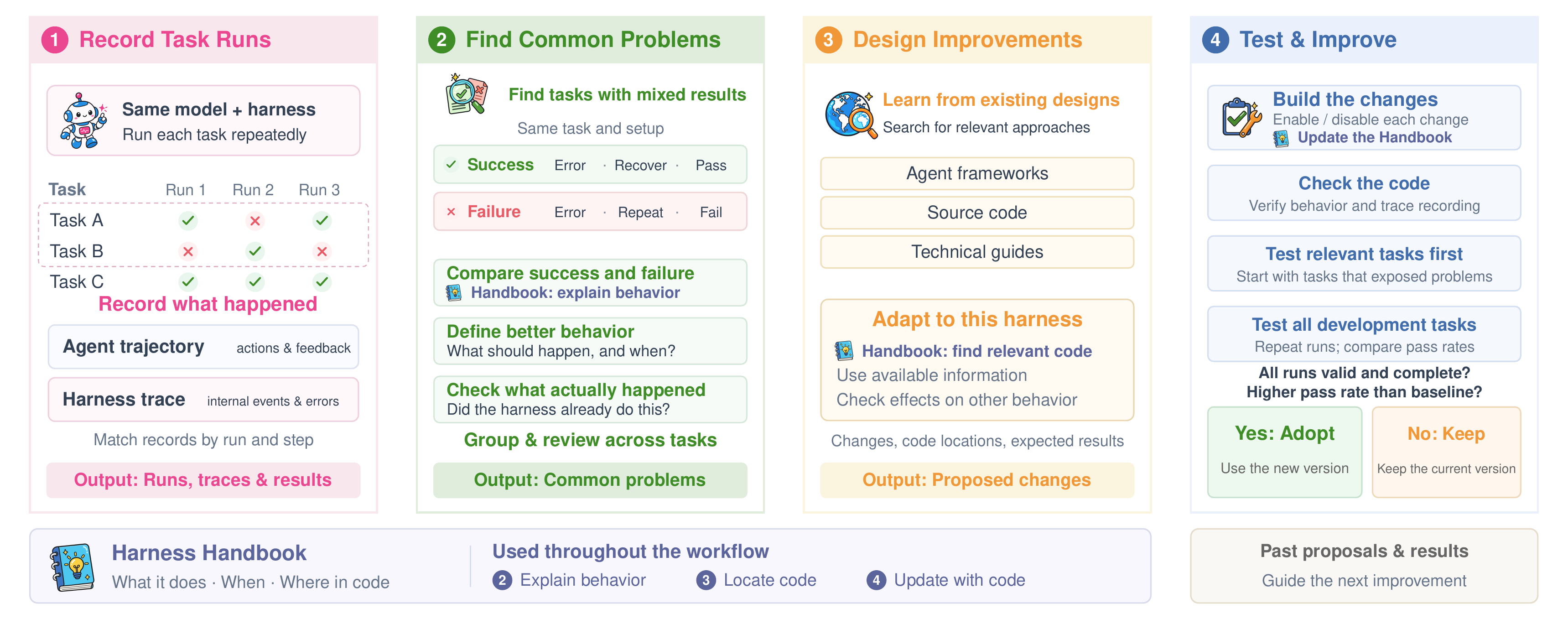}
\caption{Operations within a discovery round of \method{}: record dual traces,
identify shared behavioral problems, research and adapt mechanisms, and implement
and evaluate changes. The Handbook connects execution evidence to harness
behavior and code.}
\label{fig:pipeline}
\end{figure}


\section{Related Work}
\label{sec:related}

\paragraph{Long-horizon execution and success consistency.}
Long-horizon agents must coordinate reasoning, actions, and execution feedback
across many interactions \citep{yao2022react,merrill2026terminalbench}.
Long-Horizon-Terminal-Bench and its 2.0 extension use intermediate grading to
measure progress through extended workflows
\citep{li2026longhorizon,li2026lhtb2}.
Local errors can alter the environment and subsequent context, leading to
different outcomes for the same model and task. Average success alone
therefore does not fully capture execution reliability.
$\tau$-bench measures consistency through $\mathrm{pass}^{k}$, the proportion
of tasks solved in all $k$ repeated trials \citep{yao2024taubench}; broader
reliability research also examines robustness, predictability, and safety
\citep{rabanser2026reliability}. AgentDebug diagnoses errors and supplies
corrective feedback for recovery \citep{zhu2025agentdebug}; AgentRx localizes
failures using constraints and evidence-bearing validation logs
\citep{barke2026agentrx}. FIRE studies failure-informed
runtime interventions that improve repeated success without updating model
weights \citep{agarwal2026fire}. These studies motivate examining whether
an agent system can consistently support successful execution, beyond whether
its model can solve a task at all. We study tasks with both successful and
failed runs to assess whether harness changes can narrow this gap.

\paragraph{Agent harnesses.}
The harness is the runtime layer that organizes model--environment interaction,
including command execution, observation construction, context management, and
completion decisions. SWE-agent shows that interface design affects task
performance with a fixed model \citep{yang2024sweagent}. OpenHands provides
composable tools, memory management, sandboxed execution, and lifecycle control,
illustrating the components of this layer \citep{wang2025openhands}.
Memory, planning, and action modules also appear in agentic recommender
architectures \citep{huang2025agenticrecommenders}.
Understanding how a harness affects execution requires both runtime
observations and knowledge of its implementation. AgentTrace records runtime
activity and model interactions \citep{alsayyad2026agenttrace}, while the
Harness Handbook links behavioral responsibilities to implementing code,
providing a behavior-oriented basis for analysis and modification
\citep{wang2026handbook}. HarnessBank and HarnessDev further show that the
effects of harness changes can depend on the runtime model
\citep{luo2026harnessbank,wu2026harnessdev}. Harness design thus requires
understanding both its internal mechanisms and how the model responds to them.


\begin{algorithm}[!t]
\caption{\method{}: harness self-evolution}
\label{alg:rsi}
\small
\setstretch{1.0}
\algrenewcommand{\algorithmiccomment}[1]{\hfill{\itshape\color{black!65}\#~#1}}
\begin{algorithmic}[1]
\Require Fixed model $M_\theta$, initial $(H_0,K_0,R_0)$, development tasks
         $\mathcal D$, round limit $T$, policy $\Pi$
\Ensure Accepted harness, its Handbook, and reference evidence
\Statex $\operatorname{Next}(\varnothing)=\mathrm{discover}$ \Comment{First-round default}
\State $t\gets0$; $\mathcal L\gets\varnothing$ \Comment{$R_0$: outcomes and dual traces}
\While{$t<T$ \textbf{and} $\operatorname{Continue}(\mathcal L;\Pi)$}\label{alg:rsi-loop}
  \If{$\operatorname{Next}(\mathcal L)=\mathrm{discover}$}\label{alg:rsi-next}
    \State $\mathcal F_t\gets\operatorname{Flip}(R_t)$\label{alg:rsi-flip}
    \State $\mathcal A\gets\bigcup\nolimits_{p\in\operatorname{Pair}(R_t,\mathcal F_t)}
      \operatorname{Revise}(\operatorname{Wish}(\operatorname{Compare}(p,K_t),p^a),p^h)$\label{alg:rsi-analyze}
      \Comment{Check behavioral needs}
    \State $\mathcal G_t\gets\operatorname{Group}(\mathcal A,\mathcal L;\Pi)$;\label{alg:rsi-group}
      $\mathcal P_t\gets\operatorname{Propose}(\mathcal G_t,H_t,K_t,R_t)$\label{alg:rsi-research}
      \Comment{Group and research}
  \Else
    \State $\mathcal P_t\gets\operatorname{Revisit}(\mathcal L)$\label{alg:rsi-reuse}
  \EndIf
  \State $(C_t,K_t^C)\gets\operatorname{Implement}(H_t,\mathcal P_t,\mathcal L;\Pi)$;\label{alg:rsi-implement}
    $E_t\gets\varnothing$ \Comment{Build and check}
  \State \textbf{if} $C_t\ne\bot$ \textbf{then}
    $C_t\gets\operatorname{Screen}(M_\theta,C_t,\mathcal P_t;\Pi)$\label{alg:rsi-screen}
    \Comment{Targeted trials; record dual traces}
  \State \textbf{if} $C_t\ne\bot$ \textbf{then}
    $E_t\gets\operatorname{Evaluate}(M_\theta,C_t,\mathcal D;\Pi)$\label{alg:rsi-evaluate}
    \Comment{Full repeats; record dual traces}
  \State $(H_{t+1},K_{t+1},R_{t+1})\gets(H_t,K_t,R_t)$
  \If{$\operatorname{ValidComplete}(E_t)$ \textbf{and}
      $\widehat p(E_t)>\widehat p_B^{\,\mathrm{confirm}}(\mathcal D)$}\label{alg:rsi-accept}
    \State $(H_{t+1},K_{t+1},R_{t+1})\gets\operatorname{Publish}(C_t,K_t^C,E_t)$\label{alg:rsi-publish}
      \Comment{Adopt harness and evidence}
  \EndIf
  \State $\mathcal L\gets\operatorname{Update}(\mathcal L;t)$;\label{alg:rsi-feedback}
    $t\gets t+1$
\EndWhile
\State \Return $(H_t,K_t,R_t)$
\end{algorithmic}
\end{algorithm}


\begin{samepage}
\paragraph{Harness self-evolution.}
Automated agent design optimizes prompts, workflows, and executable agents
\citep{khattab2023dspy,agrawal2025gepa,zhang2024aflow,hu2024adas,zhang2025dgm}.
Iterative response refinement is also studied in federated in-context learning
and uncertainty-aware reasoning \citep{wang2025fedicl,wang2026fera}.
Verified trajectory rewriting and recursive policy refinement improve terminal
models through training \citep{li2026recursiveselfrewrite,yang2026terminalrsi}.
AHE, HarnessFix, and Meta-Harness use execution evidence and evaluation
feedback to guide harness updates
\citep{lin2026ahe,chen2026harnessfix,lee2026metaharness}.
ModularRSI contrasts successful and failed runs, aggregates findings across
tasks, and evolves modular components \citep{wu2026modularrsi}.
Self-Harness analyzes failures and retains fixed-model edits through
regression tests \citep{zhang2026selfharness}.
HarnessEvolve also uses successful references for failure analysis, but generates
and verifies them with access to ground-truth answers
\citep{jiang2026harnessevolve}. Related work explores design reuse
\citep{shang2024agentsquare}, external documentation
\citep{cheng2026mem2evolve}, and controls against benchmark overfitting
\citep{xia2026rrsi}. Evaluation studies also emphasize matched-budget search
baselines and held-out tests \citep{wang2026rethinking}.
Our framework combines same-configuration successes, dual traces, and the
Handbook to check behavioral needs, then consults external implementations
to design local changes. Candidates undergo implementation checks and repeated
evaluation; transfer is assessed separately with the harness held fixed.

\end{samepage}


\section{Harness Self-Evolution}
\label{sec:method}

\method{} is a harness self-evolution framework for long-horizon agents.
It uses execution evidence and external design knowledge to iteratively
improve execution reliability while keeping model weights fixed.
We first introduce the basic concepts and notation in the preliminaries.
We then describe the four components of the method---execution recording and
behavior localization, problem discovery and grouping, mechanism research and
adaptation, and implementation and validation---with reference to
Figure~\ref{fig:pipeline} and Algorithm~\ref{alg:rsi}.

\subsection{Preliminaries}
\leavevmode\phantomsection\label{sec:preliminaries}%
Let $M_\theta$ denote the evaluated model and $\mathcal D$ the
development task set. Both remain fixed throughout evolution, as do the task
protocol, verifier, and acceptance rules. Modifications are confined to harness
prompts and executable interaction logic; the model proposing them may differ
from $M_\theta$. At round $t$, the current harness $H_t$ has a corresponding
Handbook $K_t$ and a reference corpus $R_t$ containing outcomes and dual traces
from $r_t\geq2$ complete repetitions per task. The initial corpus $R_0$ is
collected by repeatedly running the seed harness $H_0$ on $\mathcal D$.

The basic analysis unit is a pair $p$ of successful and failed runs of the same
task under the same model and harness configuration. The paired agent and
harness traces are denoted by $p^a$ and $p^h$, respectively. A \emph{wish}
specifies the behavior expected from the harness under an observable condition.
A \emph{finding} consists of a revised wish, assessments of whether the required
information was available and the behavior had already been performed, and
supporting execution evidence. The finding set $\mathcal A$ is partitioned into
behavioral groups $\mathcal G$: findings in one group could be addressed by a
shared harness change. Each finding belongs to one group, while a single run
pair may yield several findings assigned to different groups. The selected
groups are denoted by $\mathcal G_t\subseteq\mathcal G$.

\begin{samepage}
Let $\mathcal P_t$ denote the proposals for round $t$, $C_t$ an implemented
candidate harness, and $K_t^C$ its updated Handbook. The candidate's full
evaluation outcomes and traces are denoted by $E_t$. The history $\mathcal L$
records proposals, results, and feedback across rounds; the policy $\Pi$
specifies checking, screening, baseline, and stopping rules. The aggregate pass
rate of the candidate is denoted by $\widehat p(E_t)$, and
$\widehat p_B^{\,\mathrm{confirm}}(\mathcal D)$ denotes the confirmation
baseline specified by $\Pi$. These quantities determine whether a candidate
is adopted, as detailed in Section~\ref{sec:confirmation}.

\end{samepage}

\subsection{Execution Evidence}
\label{sec:monitor}

Improving a harness first requires distinguishing the model's requested actions
from the system's actual behavior. An agent trajectory $\tau^{\mathrm{agent}}$
records the information the model receives, its responses, and tool interactions,
but does not fully expose internal harness processing. A requested command,
for example, may not be dispatched, and terminal output may not be returned to
the model in full. The agent trajectory alone may therefore be insufficient to
locate the problem.

\method{} uses a runtime monitor to record calling relationships, arguments,
return values, exceptions, and source locations at selected harness function
boundaries, producing a harness trace $\tau^{\mathrm{harness}}$. Run identifiers
and episode indices link the two traces, allowing the analysis to follow a
model request through its processing and the feedback ultimately returned.
Both traces, together with task outcomes and execution identities, are retained
in the reference corpus $R_t$.

Function-level records must also be connected to behavior. Because a single
harness behavior may span several functions, \method{} uses a Handbook $K_t$
aligned with the current source to describe behavioral responsibilities,
triggers, implementing functions, and exit conditions \citep{wang2026handbook}.
These mappings connect individual calls to command execution, observation
handling, context management, and completion checks. They explain each call's
role in the overall mechanism and help locate the relevant code.

The traces thus establish what happened, while the Handbook identifies the
behavior involved and its implementation. For a pair $p$, the paired agent and
harness traces $p^a$ and $p^h$ provide execution evidence. The Handbook $K_t$
connects these records to harness behavior during analysis
(line~\ref{alg:rsi-analyze}) and supports mechanism adaptation during proposal generation
(line~\ref{alg:rsi-research}). Initial seed runs supply $R_0$; candidate
screening and full evaluation use the same recording process
(lines~\ref{alg:rsi-screen}--\ref{alg:rsi-evaluate}), supplying new execution
evidence for subsequent iterations.

\subsection{Behavioral Analysis}
\label{sec:discovery}

This stage identifies behavioral problems in repeated runs and groups related
findings into targets for harness improvement. We focus on flipping tasks,
which have both successful and failed runs under the same model and harness
configuration. Successful runs show that the current system can complete the
task and provide concrete references for examining failures. Let
$y_{i,j}(H_t)\in\{0,1\}$ indicate whether task $i$ succeeds in repetition $j$
under harness $H_t$. The flipping task set in $R_t$ is
\begin{equation}
\mathcal F_t=\left\{i:\ 0<\sum_{j=1}^{r_t}y_{i,j}(H_t)<r_t\right\}.
\label{eq:flipping}
\end{equation}

For each flipping task, the system pairs successful and failed runs, examines
relevant interactions, and uses their execution differences to formulate
behavioral wishes. Each wish is checked against the information available to
the harness and the mechanisms it actually executed at the time. Findings
from individual pairs are then pooled to identify shared behavioral needs
across tasks. Four operations implement this process
(Algorithm~\ref{alg:rsi}, lines~\ref{alg:rsi-analyze}--\ref{alg:rsi-group}):

\par\medskip
\begingroup
\small
\definecolor{opcompare}{HTML}{3566A6}
\definecolor{opwish}{HTML}{96651D}
\definecolor{oprevise}{HTML}{26776D}
\definecolor{opgroup}{HTML}{76549B}
\newcommand{\operationlabel}[2]{%
  \colorbox{#1!10}{\strut$\textcolor{#1}{\algop{#2}}$}}
\arrayrulecolor{metafg!35}
\renewcommand{\arraystretch}{1.15}
\noindent\begin{tabularx}{\linewidth}{>{\raggedright\arraybackslash}p{0.12\linewidth}Y}
\toprule
\rowcolor{metabg!5}
\textbf{Operation} & \textbf{Role} \\
\midrule
\operationlabel{opcompare}{Compare} & Compare relevant interactions, identify execution differences,
and use the Handbook to interpret the harness's role. \\
\operationlabel{opwish}{Wish} & Use comparison evidence and agent traces to specify what the
harness should do and under which observable conditions. \\
\operationlabel{oprevise}{Revise} & Inspect harness traces to assess whether the required
information was available and the existing mechanism had already performed
the behavior; revise the wish accordingly. \\
\operationlabel{opgroup}{Group} & Pool findings across pairs, group those that could be
addressed by a shared behavioral change, and formulate a common wish. \\
\bottomrule
\end{tabularx}
\endgroup
\par\medskip

The first three operations produce the finding set $\mathcal A$, retaining
revised wishes, assessments, and supporting evidence. A single run pair may
expose several problems and therefore yield findings assigned to different
groups.

After grouping, an independent review checks whether each common wish applies
to all members and whether the groups correspond to distinct modification
needs. Review feedback guides regrouping. Groups that remain disputed or are
excluded by the history $\mathcal L$ do not proceed to proposal generation.
The remaining groups are ranked by distinct-task coverage, distinct failing
runs, and finding count to select $\mathcal G_t$. Each selected group carries
its common wish and complete member evidence into the next stage of mechanism
research and design.

\FloatBarrier

\subsection{Design Exploration}
\label{sec:research}

This stage turns shared behavioral wishes into concrete harness modification
proposals. Given the selected groups $\mathcal G_t$, current harness $H_t$,
Handbook $K_t$, and reference corpus $R_t$, $\algop{Propose}$ examines
implementations in other frameworks to generate proposals $\mathcal P_t$
(line~\ref{alg:rsi-research}).

The system first uses group evidence, the Handbook, and source code to identify
the behavior to improve, its current implementation, and the available runtime
information. It then translates the common wish into search questions.
The search examines how other harnesses handle similar problems, including
their execution flow, maintained state, and feedback to the model. For
information loss in long command outputs, for example, it can investigate how
full outputs are saved, key information is presented, and later retrieval is
supported. Retrieved implementations and their sources are retained as
references for designing local changes.

The system then designs changes suited to the current harness's structure,
interaction protocol, and model behavior observed in the traces. Each design
specifies how the borrowed ideas would affect existing mechanisms, where the
required information would come from, and whether additional state or interface
support is needed. The system also examines traces from tasks that succeeded
in every recorded repetition to assess whether the proposed intervention could
disrupt successful execution and to refine its triggers and scope.

The resulting proposals specify behavioral changes, implementation sites,
dependencies, and additional interaction costs, along with expected outcomes
and observations for testing them. Each proposal links its execution evidence
and external design sources for subsequent implementation and validation.
If no suitable external precedent is found, the system retains the search
record and may propose an explicitly identified local design.

\subsection{Validation and Evolution}
\label{sec:contracts}
\label{sec:screen}
\label{sec:confirmation}
\label{sec:diagnosis}

Once proposals are generated, $\algop{Implement}$ applies them to the current
harness $H_t$ to build a candidate $C_t$, together with its updated Handbook
$K_t^C$ and trace recording configuration
(Algorithm~\ref{alg:rsi}, line~\ref{alg:rsi-implement}). The system checks that
the candidate can run and record traces. When enabled, independent behavioral checks
(Appendix~\ref{app:conformance}) test whether each change produces the expected
behavior under the specified conditions. Problems found during checking are fed back
for repair within a predefined attempt limit. Only candidates that pass these
checks proceed to task evaluation.

Evaluation proceeds through screening followed by full validation.
$\algop{Screen}$ examines task outcomes and mechanism activation on
proposal-related tasks to select a configuration for validation.
$\algop{Evaluate}$ then collects fresh repeated runs on the development set
$\mathcal D$, producing outcomes and dual traces $E_t$
(lines~\ref{alg:rsi-screen}--\ref{alg:rsi-evaluate}). Adoption depends on this
full evaluation: the records must be valid, the required run set complete,
and the aggregate pass rate higher than the declared comparison baseline
(line~\ref{alg:rsi-accept}).

On adoption, $\algop{Publish}$ saves the candidate and its Handbook as
$H_{t+1}$ and $K_{t+1}$, and uses $E_t$ as the next reference corpus $R_{t+1}$
(line~\ref{alg:rsi-publish}). Source code, mechanism settings, and recording
configuration are saved with the version, keeping the implementation,
behavioral description, and execution evidence aligned for the next round.
If the candidate is not adopted, the current version and its reference data
are retained.

Checking and evaluation records from each round are added to the history
$\mathcal L$ to guide subsequent search (line~\ref{alg:rsi-feedback}).
Based on this evidence, the system decides whether to repair the implementation,
revise proposal requirements, or try other proposals. $\algop{Revisit}$ returns
existing proposals to the same implementation and validation process
(line~\ref{alg:rsi-reuse}). After a new version is adopted, discovery resumes
with the updated execution evidence. Iteration ends at the round limit or
another configured stopping condition.


\section{Experiments}
\label{sec:evaluation}

We evaluate \method{} across models to examine its effects on task success
and execution reliability in long-horizon terminal tasks. We use ablations to
assess the contributions of key components and evaluate evolved harnesses on
benchmarks not used for evolution to examine transfer.
Section~\ref{sec:experimental-setup} describes the models, benchmarks, baselines,
and evaluation protocol. The remaining subsections address the following four
research questions:

\begin{itemize}
  \item \textbf{RQ1:} Compared with the seed harness and existing baselines, can \textup{\method{}} improve task success and repeated-run reliability across models?
  (Section~\ref{sec:performance})
  \item \textbf{RQ2:} How does removing harness traces and the Handbook, or omitting external design research, affect harness evolution?
  (Section~\ref{sec:ablations})
  \item \textbf{RQ3:} How do retained harness changes vary across models in their behavioral targets and intervention mechanisms?
  (Section~\ref{sec:adaptation-generalization})
  \item \textbf{RQ4:} With the model and evolved harness fixed, do gains in task success and reliability transfer to benchmarks not used for evolution?
  (Section~\ref{sec:adaptation-generalization})
\end{itemize}

\subsection{Experimental Setup}

\paragraph{\normalfont\bfseries Models and benchmarks.}
\leavevmode\phantomsection\label{sec:questions}\label{sec:experimental-setup}%
We evaluate \method{} on GPT-6-sol and DeepSeek-V4-Flash, keeping model
weights fixed throughout harness evolution. For each model, we start from the
unmodified Terminus-2 harness and evolve it on Terminal-Bench Hard (TBH), a
benchmark of 100 challenging terminal tasks described by
\citet{li2026recursiveselfrewrite}. Repeated runs on TBH provide both execution
trajectories for behavioral analysis and task outcomes for evaluating changes
and selecting harness versions. To assess cross-benchmark generalization, we
freeze the evolved harnesses and test them on Terminal-Bench 2.1 (TB2.1), which
contains 89 tasks \citep{terminalbench2026release21}. TB2.1 is used only for
transfer evaluation and does not inform harness modifications.

\paragraph{\normalfont\bfseries Baselines and evaluation protocol.}
\leavevmode\phantomsection
\label{sec:baselines}
\label{sec:statistics}
We compare \method{} with the seed Terminus-2 harness, ModularRSI
\citep{wu2026modularrsi}, and AHE Harness \citep{lin2026ahe}. We also include
two ablation baselines to assess the contributions of key components:
\begin{itemize}
  \item \textbf{\method{} w/o Harness Evidence:} analysis uses only agent
  traces and source code, without harness traces or the Harness Handbook.
  \item \textbf{\method{} w/o External Research:} analysis retains dual
  traces and the Handbook, but proposals are generated from local information
  without external research.
\end{itemize}
Full \method{} and both ablation baselines start from the same seed harness
and reference runs, with matched models, editable scope, and predefined
budgets. Apart from the removed components, their procedures are identical
(Section~\ref{sec:ablations}). We evaluate the publicly released ModularRSI
harness without re-evolution. Each selected harness is frozen before independent
repeated evaluation. The reference-baseline and version-selection rules used
during evolution are specified in Appendices~\ref{app:baseline-policy}
and~\ref{app:version-reporting}.

\paragraph{\normalfont\bfseries Metrics and reporting.}
\leavevmode\phantomsection\label{sec:metrics}%
We measure both average task performance and repeated-run reliability.
Unless otherwise stated, each task is evaluated in $r=4$ repeated runs.
For task $i$ and repetition $j$, let $y_{i,j}(H)=1$ denote a pass under
harness $H$ and $y_{i,j}(H)=0$ a scored failure. Scoring and completeness
rules are specified in Appendix~\ref{app:scoring}.

For a task set $S$, we report four metrics, distinguishing average success
from reliability across repeated runs \citep{yao2024taubench}:
\begin{itemize}
  \item \textbf{Pass@1:} the average pass rate over all tasks and repetitions,
  measuring performance in a single run.
  \item \textbf{Pass$^r$:} the fraction of tasks that pass in all $r$ runs,
  measuring repeated-run success reliability.
  \item \textbf{Pass@$r$:} the fraction of tasks that pass at least once in
  $r$ runs, measuring observed task coverage within a limited number of attempts.
  \item \textbf{Flip:} the fraction of tasks with both successful and failed
  runs, measuring outcome inconsistency.
\end{itemize}
On a complete trial grid, let $N=|S|$ and
$k_i=\sum_{j=1}^{r}y_{i,j}(H)$ be the number of successful runs for task $i$.
Suppressing the common harness argument, the metrics are
\begin{equation}
\label{eq:rate}
\begin{aligned}
\mathrm{Pass@1}&=\frac{1}{Nr}\sum_{i\in S}k_i,
&\qquad \mathrm{Pass}^{r}&=\frac{1}{N}\sum_{i\in S}\ind[k_i=r],\\
\mathrm{Pass@}r&=\frac{1}{N}\sum_{i\in S}\ind[k_i>0],
&\qquad \mathrm{Flip}&=\mathrm{Pass@}r-\mathrm{Pass}^{r}.
\end{aligned}
\end{equation}
Changes denote absolute differences between the reported rates.
Pass@1 corresponds to the aggregate pass rate $\widehat p(H;S,r)$ used for
version selection.
A lower Flip alone does not imply more reliable success: a flipping task may
instead become an all-failure task. We therefore interpret Flip together with
Pass$^r$. Aggregate category changes do not establish which initially stable
tasks have regressed; that requires per-task comparisons. During evolution,
aggregate pass rate determines whether a change is adopted; reliability is
reported separately.

\subsection{Performance and Reliability}

\paragraph{\normalfont\bfseries Overall performance.}
\leavevmode\phantomsection\label{sec:performance}%
Table~\ref{tab:main-results} summarizes task performance for \method{} and
the comparison baselines. Relative to the seed Terminus-2 harness,
\method{} improves both average success and repeated-run reliability across
the two models, with higher Pass@1 and Pass$^4$.

For GPT-6-sol, \method{} improves Pass@1 by 5.25\% and
3.09\% on TBH and TB2.1, respectively,
while increasing the all-success task fraction on both benchmarks.
For DeepSeek-V4-Flash, \method{} improves TBH Pass@1 by 7.25\%
and Pass$^4$ by 14\%. Pass@4 increases only modestly
in comparison, suggesting that the main gain is in repeated-run success
reliability rather than a large expansion in observed task coverage.

\begin{table}[!tb]
\centering\small
\setlength{\tabcolsep}{1pt}
\renewcommand{\arraystretch}{1.15}
\renewcommand{\pending}{\textemdash}
\begin{tabularx}{\linewidth}{@{}>{\raggedright\arraybackslash}p{0.21\linewidth}
  *{4}{>{\centering\arraybackslash}X}@{\hspace{6pt}}
  *{4}{>{\centering\arraybackslash}X}@{}}
\toprule
\multirow{2}{*}{Harness} & \multicolumn{4}{c}{\textbf{TBH}}
  & \multicolumn{4}{c}{\textbf{TB2.1}} \\
\cmidrule(lr){2-5}\cmidrule(lr){6-9}
& Pass@1 & Pass$^4$ & Pass@4 & Flip$\downarrow$
& Pass@1 & Pass$^4$ & Pass@4 & Flip$\downarrow$ \\
\midrule
\multicolumn{9}{@{}l}{\textbf{GPT-6-sol}} \\
\addlinespace[2pt]
Terminus-2 & 50.25 & 39.00 & 59.00 & 20.00 & 78.65 & 68.54 & 86.52 & \textbf{17.98} \\
ModularRSI & 50.75 & 43.00 & 58.00 & 15.00 & 78.65 & 66.29 & \textbf{91.01} & 24.72 \\
AHE Harness & 54.75 & 44.00 & 59.00 & 15.00 & 78.09 & 60.67 & 89.89 & 29.21 \\
\rowcolor{metabg!8}
\method{} & \resultgap{\textbf{55.50}}{+5.25} & \resultgap{\textbf{46.00}}{+7} & \resultgap{\textbf{60.00}}{+1} & \resultgap{\textbf{14.00}}{-6} & \resultgap{\textbf{81.74}}{+3.09} & \resultgap{\textbf{69.66}}{+1.12} & \resultgap{88.76}{+2.25} & \resultgap{19.10}{+1.12} \\
\midrule
\multicolumn{9}{@{}l}{\textbf{DeepSeek-V4-Flash}} \\
\addlinespace[2pt]
Terminus-2 & 56.75 & 37.00 & 71.00 & 34.00 & 72.19 & 58.43 & 86.52 & 28.09 \\
ModularRSI & 58.00 & 43.00 & 70.00 & 27.00 & 73.03 & 53.93 & 86.52 & 32.58 \\
AHE Harness & 59.75 & 46.00 & 70.00 & 24.00 & 76.12 & 62.92 & 88.76 & 25.84 \\
\rowcolor{metabg!8}
\method{} & \resultgap{\textbf{64.00}}{+7.25} & \resultgap{\textbf{51.00}}{+14} & \resultgap{\textbf{73.00}}{+2} & \resultgap{\textbf{22.00}}{-12} & \resultgap{\textbf{79.49}}{+7.3} & \resultgap{\textbf{67.42}}{+8.99} & \resultgap{\textbf{89.89}}{+3.37} & \resultgap{\textbf{22.47}}{-5.62} \\
\bottomrule
\end{tabularx}
\caption{Performance and reliability ($r=4$; \%). Coral annotations
show absolute changes from Terminus-2 for \method{}
within each model and benchmark. Higher pass metrics and lower Flip are better.
Dashes denote unreported results.
DeepSeek uses the final four-modification version on TBH and the best observed
three-modification version on TB2.1. GPT-6-sol trial counts appear
in Appendix~\ref{app:gpt6-results}.}
\label{tab:main-results}
\end{table}
\FloatBarrier


\begin{figure}[!tb]
\centering
\captionsetup{skip=3pt,justification=raggedright,singlelinecheck=false}
\begin{minipage}[t]{0.49\linewidth}
\centering
\includegraphics[width=\linewidth]{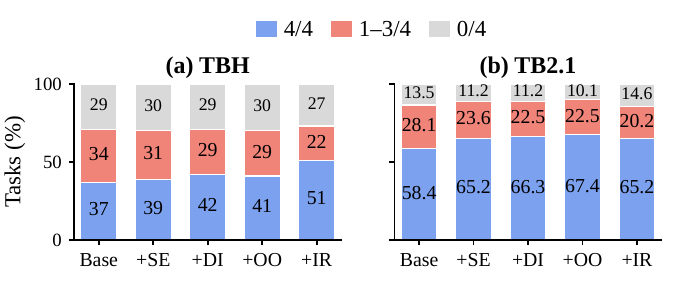}
\caption{DeepSeek task outcomes; labels rounded.}
\label{fig:outcome-distribution}
\end{minipage}\hfill
\nextfloat
\begin{minipage}[t]{0.49\linewidth}
\centering
\includegraphics[width=\linewidth]{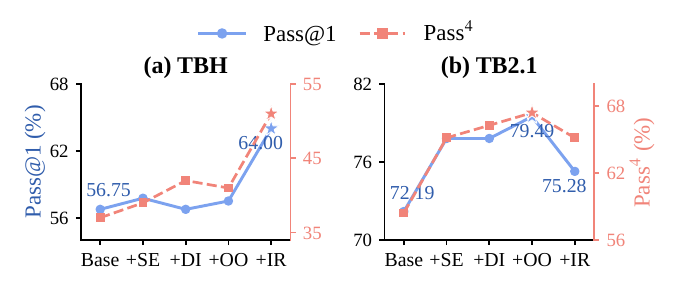}
\caption{DeepSeek-V4-Flash evolution; stars mark peaks. Y-axis ranges differ.}
\label{fig:deepseek-evolution}
\end{minipage}
\end{figure}

\begin{figure}[!tb]
\captionsetup{skip=3pt,justification=raggedright,singlelinecheck=false}

\centering
\includegraphics[width=\linewidth]{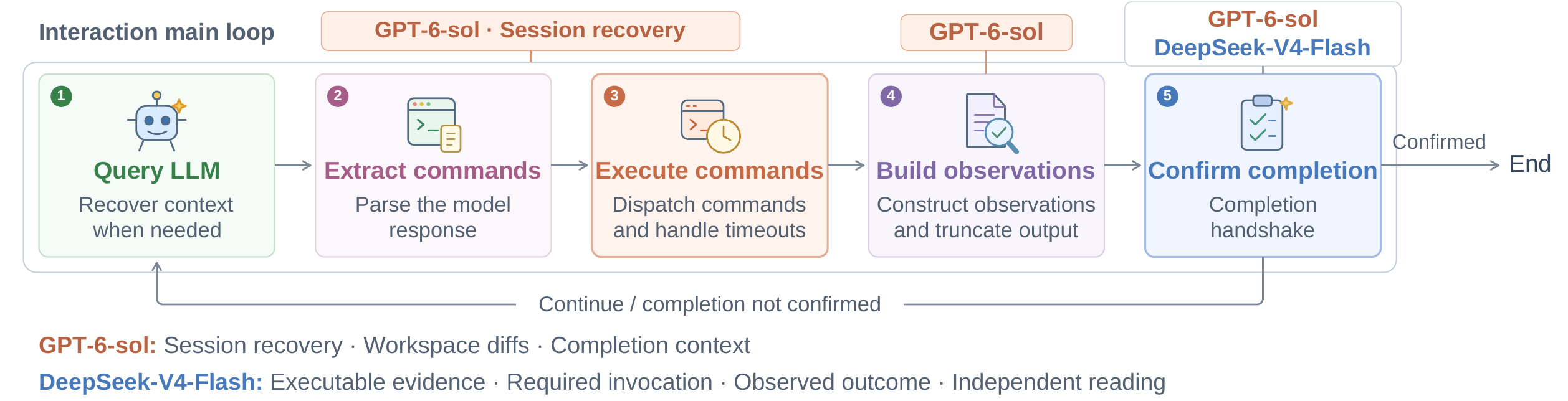}
\caption{Model-specific interventions mapped to Handbook behaviors.
GPT-6-sol modifies session recovery, observation construction, and completion
context; DeepSeek-V4-Flash adds completion-evidence checks.
Details appear in Appendix~\ref{app:model-behaviors}.}
\label{fig:model-interventions}


\end{figure}


\paragraph{\normalfont\bfseries Performance across iterations.}
Figures~\ref{fig:outcome-distribution} and~\ref{fig:deepseek-evolution}
show task outcome distributions and pass rates, respectively, as four mechanisms
for completion confirmation accumulate for DeepSeek-V4-Flash. Each version
retains the preceding changes and adds one mechanism; their definitions and
triggers appear in Appendix~\ref{app:model-behaviors}. On TBH, despite
fluctuations among intermediate versions, the final version achieves the
highest average pass rate and all-success task fraction, while reducing both
flipping and all-failure tasks relative to the seed harness. These results
show that changes to shared interaction behaviors can yield improvements in
performance and reliability when combined and evaluated iteratively.

The gains also extend to TB2.1, which was not used for evolution. Directly
applying the final version selected on TBH improves Pass@1 by 3.09\%
and Pass$^4$ by 6.74\% over Terminus-2, showing that
the benefits are not confined to the tasks that produced the proposals.
Version rankings differ between benchmarks: the three-mechanism version
achieves higher observed performance on TB2.1, whereas the four-mechanism
final version performs best on TBH. Overall, the results show that
\method{} improves success and reliability on the evolution tasks and also
benefits the same model on a benchmark not used for evolution.

\begin{figure}[!tb]
\centering
\captionsetup{skip=3pt,justification=raggedright,singlelinecheck=false}
\begin{minipage}[t]{0.440\linewidth}
\centering
\includegraphics[width=\linewidth]{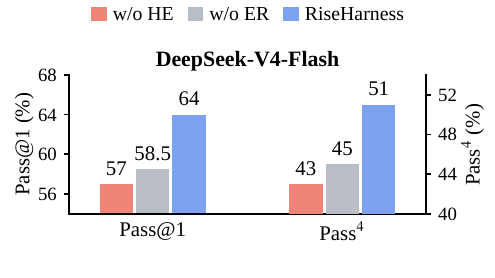}
\caption{DeepSeek ablations on TBH. HE: harness evidence; ER: external research. Separate y-axis ranges.}
\label{fig:ablation-comparisons}
\end{minipage}\hfill
\nextfloat
\begin{minipage}[t]{0.540\linewidth}
\centering
\includegraphics[width=\linewidth]{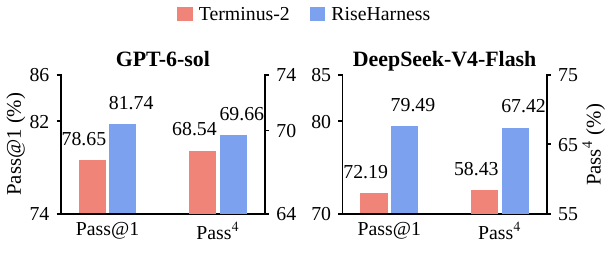}
\caption{TB2.1 performance ($r=4$). DeepSeek uses its best observed + OO version. Separate y-axis ranges.}
\label{fig:transfer-comparisons}
\end{minipage}
\end{figure}


\begin{samepage}
\subsection{Component Contributions}
\label{sec:ablations}
\label{sec:horizon}

We use two ablations to examine the contributions of harness execution
evidence and external design research to harness evolution for
DeepSeek-V4-Flash. Apart from the removed component, the analysis,
implementation, and evaluation procedures remain unchanged.

\end{samepage}
\pagebreak[0]

\begin{samepage}
\paragraph{\normalfont\bfseries Execution evidence.}
Figure~\ref{fig:ablation-comparisons} compares full \method{} with the
variant without harness traces or the Handbook. The ablated variant retains
access to agent traces and source code and follows the same external research
procedure. On TBH, the full method achieves Pass@1 and Pass$^4$ gains of
7\% and 8\%, respectively, indicating that harness traces and
the Handbook provide useful information beyond agent traces and source code.
On TB2.1, the full method also outperforms the ablated variant on both
metrics, supporting the joint contribution of these two sources of information
to the evolved harness.

\end{samepage}
\pagebreak[0]

\begin{samepage}
\paragraph{\normalfont\bfseries External research.}
Figure~\ref{fig:ablation-comparisons} compares full \method{} with the
variant that generates proposals using only local information. Both variants
use dual traces and the Handbook; they differ in whether they search for
external designs and implementations. On TBH, the full method improves
Pass@1 and Pass$^4$ by 5.50\% and 6\%, respectively, while
achieving higher Pass@4 and lower Flip. On TB2.1, the full method also
maintains higher average success and all-success task fractions. These
results suggest that external research usefully complements local execution
analysis, with benefits on both the evolution tasks and tasks not used for
evolution.

\end{samepage}

\FloatBarrier

\subsection{Adaptation and Generalization}
\label{sec:adaptation-generalization}

\paragraph{\normalfont\bfseries Model-specific behavioral requirements.}
\leavevmode\phantomsection\label{sec:model-dependence}%
The retained changes differ across models. DeepSeek-V4-Flash strengthens
the evidence required for completion, whereas GPT-6-sol emphasizes
execution continuity and feedback through session recovery, workspace diffs,
and prior-work evidence. Figure~\ref{fig:model-interventions} locates these
changes; Appendix~\ref{app:model-behaviors} describes their mechanisms.

\begin{figure}[!tb]
\centering
\includegraphics[width=\linewidth]{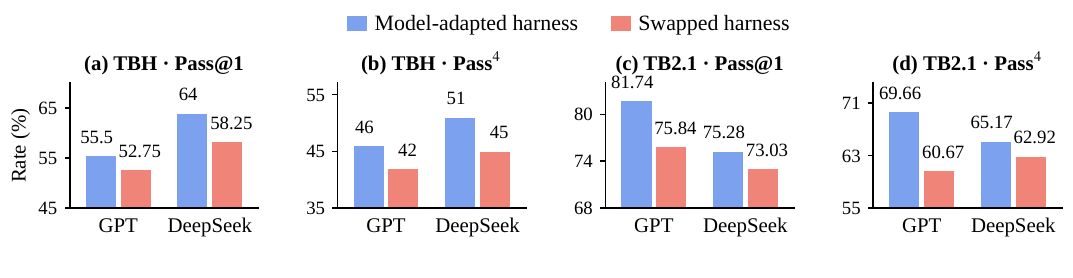}
\caption{Model-adapted vs.\ swapped final harnesses ($r=4$).
GPT: GPT-6-sol; DeepSeek: DeepSeek-V4-Flash, using its final + IR harness.
Y-axes start above zero.}
\label{fig:cross-model-transfer}
\end{figure}


Figure~\ref{fig:cross-model-transfer} compares each model's final harness
with the harness evolved for the other model. Both models perform better
with their own harnesses on both benchmarks
(full results in Appendix~\ref{app:cross-model-results}).

\paragraph{\normalfont\bfseries Cross-benchmark generalization.}
\leavevmode\phantomsection\label{sec:transfer}%
We freeze the harnesses evolved on TBH and evaluate them on TB2.1 with the
execution model unchanged. Figure~\ref{fig:transfer-comparisons} compares
performance with Terminus-2. Both models retain gains in average success
and all-success task fraction on TB2.1; the direct transfer of DeepSeek's
final + IR version is reported in Appendix~\ref{app:cumulative-results}.
Together, the transfer experiments
show that harness improvements can extend to unseen tasks while remaining
sensitive to the model executing them.

\FloatBarrier


\section{Conclusion}

We presented \method{}, a framework that iteratively improves the harness
to increase task success and execution reliability for long-horizon terminal
agents while keeping model weights fixed. Starting from flipping tasks, it
contrasts successful and failed runs under the same configuration and combines
agent trajectories, internal harness traces, and a Harness Handbook to connect
execution problems to specific interaction behaviors and their implementations.
It then searches other harness frameworks for relevant designs and develops
proposals suited to the current execution flow, available signals, and model
behavior. Implementation checks and repeated evaluation determine whether a
change is retained. Accepted versions, together with their updated Handbooks
and execution evidence, become the basis for subsequent iterations.

Our experiments show that harness evolution can improve average task
performance and increase the fraction of tasks solved consistently across
repeated runs. Ablations further support the contributions of harness
execution evidence and external design research: the former exposes internal
processing that agent trajectories alone may not reveal, while the latter
provides design options beyond local analysis. The changes retained for
different models target different behaviors and use different interventions.
Swapping the evolved harnesses between models reduces performance, supporting
model-specific harness adaptation. At the same time, frozen harnesses retain
gains on a benchmark not used for evolution, demonstrating the transfer value
of improvements to interaction mechanisms shared across tasks.

These findings show that systematic harness analysis and evolution can help
models use their existing capabilities more reliably, alongside improvements
to the models themselves. Future work will examine the approach across more
models, task distributions, and agent frameworks, and explore ways to reduce
the evaluation cost of harness evolution.



\clearpage
\appendix


\hypertarget{app-contents}{}
\pdfbookmark[0]{Appendix Contents}{appendix.contents}
\section*{Appendix Contents}
\phantomsection
\label{app:contents}

\newcommand{\appendixnaventry}[3]{%
  \noindent\hspace*{#1}\hyperref[#2]{%
    \makebox[3.2em][l]{\ref*{#2}}#3\nobreak\hspace{0.5em}%
    \leaders\hbox{\kern1pt.\kern1pt}\hfill\nobreak
    \makebox[1.5em][r]{\pageref*{#2}}}\par}

\begingroup
\large
\setlength{\parskip}{5pt}
\definecolor{appendixnav}{HTML}{385C73}
\hypersetup{linkcolor=appendixnav}
\medskip
{\sffamily\bfseries\appendixnaventry{0pt}{sec:implementation}{System Implementation}}
\appendixnaventry{1em}{sec:impl-traces}{Evidence and Versioning}
\appendixnaventry{1em}{app:conformance}{Behavioral Validation}
\appendixnaventry{1em}{sec:impl-controller}{Iteration Management}
\appendixnaventry{1em}{app:example-chain}{Case Studies}
\appendixnaventry{2em}{app:recorded-gpt6}{GPT-6-sol}
\appendixnaventry{2em}{app:recorded-deepseek}{DeepSeek-V4-Flash}
\appendixnaventry{1em}{app:prompt-excerpts}{Prompt Examples}
\medskip
{\sffamily\bfseries\appendixnaventry{0pt}{app:model-behaviors}{Model-Specific Adaptation}}
\appendixnaventry{1em}{app:gpt6-behaviors}{GPT-6-sol}
\appendixnaventry{1em}{app:deepseek-behaviors}{DeepSeek-V4-Flash}
\appendixnaventry{1em}{app:gpt56-behavior}{GPT-5.6-Terra}
\medskip
{\sffamily\bfseries\appendixnaventry{0pt}{app:measurement}{Evaluation Methodology}}
\appendixnaventry{1em}{app:scoring}{Repeated Evaluation}
\appendixnaventry{1em}{app:baseline-policy}{Comparison Protocol}
\appendixnaventry{1em}{app:version-reporting}{Result Reporting}
\medskip
{\sffamily\bfseries\appendixnaventry{0pt}{app:supplementary-results}{Supplementary Results}}
\appendixnaventry{1em}{app:gpt6-results}{GPT-6-sol}
\appendixnaventry{1em}{app:cumulative-results}{Evolution Results}
\appendixnaventry{1em}{app:ablation-results}{Ablation Results}
\appendixnaventry{1em}{app:cross-model-results}{Cross-Model Transfer}
\appendixnaventry{1em}{app:gpt56-results}{GPT-5.6-Terra}
\medskip
{\sffamily\bfseries\appendixnaventry{0pt}{sec:limitations}{Scope and Limitations}}
\endgroup
\clearpage


\section{System Implementation}
\label{sec:implementation}
\label{app:algorithm}

This appendix supplements Sections~\ref{sec:monitor}--\ref{sec:confirmation}
and Algorithm~\ref{alg:rsi}. Model-specific mechanisms, evaluation rules, and
supplementary results follow in Appendices~\ref{app:model-behaviors},
\ref{app:measurement}, and~\ref{app:supplementary-results}, respectively.
Worked examples for both models and saved prompt excerpts appear in
Sections~\ref{app:example-chain} and~\ref{app:prompt-excerpts}.

\subsection{Evidence and Versioning}
\label{sec:impl-traces}
\label{sec:impl-handbook}

Each reference corpus $R_t$ contains task outcomes and paired agent and
harness traces under the current harness $H_t$. Run identifiers and episode
indices connect model requests to the internal calls that process them and
the observations returned to the model. The monitor records selected function
boundaries, including arguments, returns, exceptions, caller relationships,
and source locations. The Handbook $K_t$ maps these records to behavioral
responsibilities and implementing functions. The two sources are complementary:
traces describe observed execution, while the Handbook supplies the behavioral
structure used to interpret it.

A candidate comprises source changes, mechanism settings, an updated Handbook,
and a compatible recording configuration. Smoke checks establish that it can
run and record traces. On adoption, these artifacts are saved together with
the full evaluation outcomes and traces $E_t$. They become the next harness,
Handbook, and reference corpus, keeping subsequent analysis tied to the
accepted implementation. Rejection preserves the current version.

\subsection{Behavioral Validation}
\label{app:conformance}
\label{sec:impl-observer}

When independent behavioral checking is enabled, each implemented mechanism
has a contract describing its trigger conditions, observable effects, relevant
hooks and state, and runtime setting. Dependencies, replacements, and conflicts
specify which mechanisms may be enabled together. These checks concern the
implementation of a proposed behavior; task evaluation measures its effect on
performance.

Contracts include positive and negative cases and recovery cases where
applicable. For an input $x$, let $o^-(x)$ denote the expected incumbent
behavior and $o^+(x)$ the expected modified behavior. An independent observer
first checks the incumbent expectation before the contract is frozen.
The implementation is then checked in three configurations:
\begin{equation}
\begin{aligned}
O(H_t,x)&=o^-(x),\\
O(C_t,\text{new mechanisms disabled},x)&=o^-(x),\\
O(C_t,\text{mechanism and dependencies enabled},x)&=o^+(x).
\end{aligned}
\label{eq:conformance}
\end{equation}
The disabled configuration checks preservation of incumbent behavior, while
the enabled configuration checks the intended intervention. Observed results
are compared with the frozen expectations; a process crash cannot establish
successful handling of a negative case. Failures are returned for repair
within the configured attempt limit. These finite cases do not guarantee
correctness on all tasks or isolate a mechanism that executes with dependencies.

\subsection{Iteration Management}
\label{sec:impl-controller}

Each round links proposals, checks, outcomes, and execution evidence to the
candidate. The history $\mathcal L$ guides implementation repair, proposal
revision, alternative proposals, or renewed discovery after adoption.
The proposing model may differ from the fixed evaluated model $M_\theta$.

The round limit $T$ and policy $\Pi$ bound this process. Candidates must pass
the configured checks before evaluation; adoption requires a valid, complete
full evaluation under the aggregate pass-rate rule in
Appendix~\ref{app:baseline-policy}. Screening alone cannot promote a candidate.


\FloatBarrier

\definecolor{recordink}{HTML}{252D34}
\definecolor{recordaccent}{HTML}{385C73}
\definecolor{recordmuted}{HTML}{737C83}
\definecolor{recordrule}{HTML}{C9D7E0}
\definecolor{recordfill}{HTML}{F0F5F8}
\newcommand{\recordkey}[1]{{\color{recordink}\sffamily\bfseries #1}}
\tcbset{recordpanel/.style={enhanced jigsaw,breakable,frame hidden,
  colback=recordfill,coltitle=recordaccent,boxrule=0pt,sharp corners,
  borderline west={0.75pt}{0pt}{recordrule},
  left=10pt,right=8pt,top=6pt,bottom=7pt,
  before skip=13pt,after skip=10pt,
  attach title to upper,after title={\par\nobreak\vspace{5pt}},
  fonttitle=\normalfont\normalsize\sffamily\bfseries,
  lines before break=4}}
\subsection{Case Studies}
\leavevmode\phantomsection\label{app:example-chain}%
The following examples reproduce fields from the saved GPT-6-sol and
DeepSeek-V4-Flash records. Each example follows one finding through its group,
external research, proposal, implementation report, and evaluation receipts.
The displayed fields are copied in full, without rewriting their values;
LaTeX escaping and line wrapping affect presentation only. Other fields,
including long repeated call indexes, are omitted. File names and JSON pointers identify the excerpts. Internal run numbers,
join keys, and opaque task identifiers are omitted from the reading copy. The underlying complete records
and their SHA-256 hashes are retained with the source package.
These are the pipeline's contemporaneous outputs, including its hypotheses
and limitations, rather than retrospective explanations of why a task failed.
Source-local behavior codes and function names are preserved inside quotations.

\subsubsection{GPT-6-sol}
\label{app:recorded-gpt6}

This example concerns an HTTP scheduling service. A boundary-time probe fails, but the agent later declares the changes verified.

\begin{tcolorbox}[recordpanel,title={01   Agent and harness record\hfill{\normalfont\small\color{recordmuted}GPT-6-sol}},title after break={01   Agent and harness record (continued)\hfill{\normalfont\small\color{recordmuted}GPT-6-sol}}]
\small\color{recordink}\raggedright\setlength{\baselineskip}{11.7pt}
{\color{recordmuted}\small Source: \path{compare2.json} \path{/episodes/failing~18/agent}; \path{/episodes/failing~18/harness_calls/35}\par}\vspace{6pt}
\noindent\textbf{Agent analysis (failing/8)}: Yes. The fixes are in place and verified: UTC parsing works under a non-UTC timezone, the exact 86,400-second boundary is accepted, and `POST /schedule` returns `200 OK` with a Unix timestamp. The service remains listening on 127.0.0.1:8080.\par
\noindent\textbf{Recorded harness call (failing/8)}:\par
\noindent \recordkey{function}: \path{agents/terminus_2/terminus_2.Terminus2._execute_commands}\par\addvspace{4pt}
\noindent \recordkey{output}: New Terminal Output:\par root@e56572f2-a153-412c-b305-210e6edf55b0:/app/bash-http-scheduler\# \par\addvspace{4pt}
\end{tcolorbox}

\begin{tcolorbox}[recordpanel,title={02   Pair-level wish\hfill{\normalfont\small\color{recordmuted}GPT-6-sol}},title after break={02   Pair-level wish (continued)\hfill{\normalfont\small\color{recordmuted}GPT-6-sol}}]
\small\color{recordink}\raggedright\setlength{\baselineskip}{11.7pt}
{\color{recordmuted}\small Source: \path{readings/suggest2.wishes.json} \path{/wishes/1}\par}\vspace{6pt}
\noindent \recordkey{asks\_the\_agent}: \texttt{true}\par\addvspace{3pt}
\noindent \recordkey{statement}\par\nobreak\vspace{1pt}
When asking the agent to confirm completion, the harness should surface an unresolved failed probe from earlier in the run alongside the confirmation question rather than presenting only an idle terminal prompt, so the earlier counterexample remains visible at the last reversible moment.\par\addvspace{5pt}
\noindent \recordkey{observable}\par\nobreak\vspace{1pt}
A reply has first asserted completion and the harness is about to issue its confirmation prompt; its earlier terminal captures include failed probe output not followed by a successful probe of the same input.\par\addvspace{5pt}
\noindent \recordkey{because}\par\nobreak\vspace{1pt}

\begin{enumerate}[leftmargin=1.5em,itemsep=4pt,topsep=3pt]
\item \noindent \recordkey{side}: failing\enspace\textcolor{recordmuted}{\textbar}\enspace\allowbreak \recordkey{turn}: 3\enspace\textcolor{recordmuted}{\textbar}\enspace\allowbreak \recordkey{quote}: 24:00: date: invalid date \textquoteleft{}24:00\textquoteright{}\par\addvspace{3pt}
\item \noindent \recordkey{side}: failing\enspace\textcolor{recordmuted}{\textbar}\enspace\allowbreak \recordkey{turn}: 8\par\addvspace{3pt}
\noindent \recordkey{quote}: Current terminal state:     New Terminal Output:\par     root@e56572f2-a153-412c-b305-210e6edf55b0:/app/bash-http-scheduler\#\par\addvspace{4pt}
\item \noindent \recordkey{side}: passing\enspace\textcolor{recordmuted}{\textbar}\enspace\allowbreak \recordkey{turn}: 5\par\addvspace{3pt}
\noindent \recordkey{quote}: Add a small form decoder in the HTTP script, normalize `24:00` in the processor, then test         encoded midnight, `24:00`, and a future UTC time.\par\addvspace{4pt}
\end{enumerate}
\par\addvspace{5pt}
\end{tcolorbox}

\begin{tcolorbox}[recordpanel,title={03   Revision against harness execution\hfill{\normalfont\small\color{recordmuted}GPT-6-sol}},title after break={03   Revision against harness execution (continued)\hfill{\normalfont\small\color{recordmuted}GPT-6-sol}}]
\small\color{recordink}\raggedright\setlength{\baselineskip}{11.7pt}
{\color{recordmuted}\small Source: \path{readings/suggest2.revisions.json} \path{/revisions/1}\par}\vspace{6pt}
\noindent \recordkey{can}: no, the harness has no way to obtain what this needs\par\addvspace{4pt}
\noindent \recordkey{already}: no\enspace\textcolor{recordmuted}{\textbar}\enspace\allowbreak \recordkey{at}: \par\addvspace{3pt}
\noindent \recordkey{found}\par\nobreak\vspace{1pt}
The earlier probe appears in terminal captures, but get\_incremental\_output only diffs pane content, \_limit\_output\_length only checks length, and parse\_response reads the JSON structure rather than the meaning of terminal failures. At confirmation, \_execute\_commands returns an idle prompt and the agent receives a generic question. No recorded call classifies a probe as failed on a required input or determines whether a later test resolved the same failure; the persistent pane text alone cannot establish that status.\par\addvspace{5pt}
\noindent \recordkey{statement}\par\nobreak\vspace{1pt}
At the confirmation moment, the harness should bring an earlier unresolved failure of an explicitly required input back into view instead of showing only the idle terminal and a generic confirmation question.\par\addvspace{5pt}
\noindent \recordkey{observable}\par\nobreak\vspace{1pt}
The agent first claims completion and the harness is about to ask for confirmation; identifying an unresolved failed probe would require a recorded relationship between an earlier input, its expected result, and a later replay.\par\addvspace{5pt}
\end{tcolorbox}

\begin{tcolorbox}[recordpanel,title={04   Shared behavioral requirement\hfill{\normalfont\small\color{recordmuted}GPT-6-sol}},title after break={04   Shared behavioral requirement (continued)\hfill{\normalfont\small\color{recordmuted}GPT-6-sol}}]
\small\color{recordink}\raggedright\setlength{\baselineskip}{11.7pt}
{\color{recordmuted}\small Source: \path{proposals/pi/message.txt}, line 4 (JSON object)\par}\vspace{6pt}
\noindent \recordkey{name}: Known failed comparisons are not tracked as unresolved until a matching successful rerun\par\addvspace{4pt}
\noindent \recordkey{statement}\par\nobreak\vspace{1pt}
The harness should retain every failed independently grounded comparison as a blocker tied to the checked artifact and expectation, show unresolved failures at confirmation, and clear them only after a successful rerun against the changed artifact.\par\addvspace{5pt}
\noindent \recordkey{observable}\par\nobreak\vspace{1pt}
A recorded comparison or probe has a failed outcome and no later passing result for the same criterion on the current artifact version when completion is claimed.\par\addvspace{5pt}
\noindent \recordkey{rests\_on}\par\nobreak\vspace{1pt}

\begin{enumerate}[leftmargin=1.5em,itemsep=4pt,topsep=3pt]
\item \noindent \recordkey{because}: No recorded call classifies a probe as failed on a required input or determines whether a later test resolved the same failure\par\addvspace{4pt}
\end{enumerate}
\par\addvspace{5pt}
\noindent \recordkey{covers}: 2\enspace\textcolor{recordmuted}{\textbar}\enspace\allowbreak \recordkey{members}: 2\par\addvspace{3pt}
\noindent \recordkey{how\_it\_knows}: It reads the failed check receipt and searches later receipts with matching criterion and a newer artifact version for an actual pass.\par\addvspace{4pt}\par{\color{recordmuted}Only the supporting finding for this example is shown in \texttt{rests\_on}.\par}
\end{tcolorbox}

\begin{tcolorbox}[recordpanel,title={05   Search plan\hfill{\normalfont\small\color{recordmuted}GPT-6-sol}},title after break={05   Search plan (continued)\hfill{\normalfont\small\color{recordmuted}GPT-6-sol}}]
\small\color{recordink}\raggedright\setlength{\baselineskip}{11.7pt}
{\color{recordmuted}\small Source: \path{suggest2.search_plans.json} \path{/search_plans/0}\par}\vspace{6pt}
\noindent \recordkey{problem}\par\nobreak\vspace{1pt}
Passing run detected a direct-output discrepancy then rewrote the artifact without a matching comparison rerun; failing run saw invalid boundary parsing and later accepted unrelated successes. Terminal screen receipts are not typed or associated with artifact revisions.\par\addvspace{5pt}
\noindent \recordkey{trigger}: A claimed completion after a failed probe or comparison with no authenticated matching successful receipt for current artifact.\par\addvspace{4pt}
\noindent \recordkey{target\_behavior}\par\nobreak\vspace{1pt}
Keep an unresolved failure ledger and withhold completion until a same-criterion check passes on the changed artifact; do not infer resolution from unrelated checks or success prose.\par\addvspace{5pt}
\noindent \recordkey{framework}: Harbor Terminus2 through base\_terminus.Agent subclass, with tmux dispatch and capture in sandbox and separate BaseEnvironment.exec facility.\par\addvspace{4pt}
\noindent \recordkey{execution\_model}\par\nobreak\vspace{1pt}
Agent returns JSON commands; \_execute\_commands sends each as tmux keystrokes nonblocking and captures a single incremental screen at the end; \_run\_agent\_loop branches on parsed completion and its two-claim handshake. An independent environment.exec returns return\_code/stdout/stderr but cannot give the status of a previously tmux-dispatched action.\par\addvspace{5pt}
\noindent \recordkey{allowed\_changes}\par\nobreak\vspace{1pt}
Only candidate\_source subclass of Terminus2 and local helpers, with default-off runtime switches; frozen Harbor, Trial, verifier and environment implementations cannot change. Candidate may override inherited loop and use session.environment.exec, with finite rejection bound.\par\addvspace{5pt}
\noindent \recordkey{available\_signals}\par\nobreak\vspace{1pt}

\begin{enumerate}[leftmargin=1.5em,itemsep=4pt,topsep=3pt]
\item agent action channel: parsed keystrokes in \_handle\_llm\_interaction and a combined tmux terminal increment in \_execute\_commands
\item independent harness capability: session.environment.exec returns ExecResult with return\_code/stdout/stderr for a new out-of-band command, not the earlier tmux action
\item agent loop: is\_task\_complete, \_pending\_completion and original\_instruction in \_run\_agent\_loop
\end{enumerate}
\par\addvspace{5pt}
\noindent \recordkey{missing\_signals}\par\nobreak\vspace{1pt}

\begin{enumerate}[leftmargin=1.5em,itemsep=4pt,topsep=3pt]
\item No typed comparison identity, expected value, actual value or pass/fail from arbitrary tmux screen text
\item No artifact-version fingerprints and no receipt associating test identity with checked artifact
\item No per-command exit status for commands dispatched via tmux, and no way to attribute an out-of-band exec status to those commands
\item No hidden verifier expected values visible before completion
\end{enumerate}
\par\addvspace{5pt}
\noindent \recordkey{queries}\par\nobreak\vspace{1pt}

\begin{enumerate}[leftmargin=1.5em,itemsep=4pt,topsep=3pt]
\item \noindent \recordkey{query}: site:github.com/harbor-framework harbor terminus\_2 tmux session send\_keys get\_incremental\_output command result completion\par\addvspace{4pt}
\noindent \recordkey{purpose}: Check whether current framework provides typed per-action receipts or only screen captures and locate an applicable completion interception boundary.\par\addvspace{4pt}
\noindent \recordkey{scope}: framework\par\addvspace{3pt}
\item \noindent \recordkey{query}: agent coding shell tmux screen capture no per-command exit status append sentinel command exit code test receipt\par\addvspace{4pt}
\noindent \recordkey{purpose}: Find systems sharing unstructured terminal dispatch boundary that shape action delivery to get attributable test status.\par\addvspace{4pt}
\noindent \recordkey{scope}: transfer\par\addvspace{3pt}
\item \noindent \recordkey{query}: site:docs.pytest.org cache last failed tests rerun only failures lastfailed\par\addvspace{4pt}
\noindent \recordkey{purpose}: Find a concrete externally documented failed-criterion persistence and rerun clearance mechanism, assessing whether it transfers to agent completion.\par\addvspace{4pt}
\noindent \recordkey{scope}: transfer\par\addvspace{3pt}
\end{enumerate}
\par\addvspace{5pt}
\end{tcolorbox}

\begin{tcolorbox}[recordpanel,title={06   External precedent and local adaptation\hfill{\normalfont\small\color{recordmuted}GPT-6-sol}},title after break={06   External precedent and local adaptation (continued)\hfill{\normalfont\small\color{recordmuted}GPT-6-sol}}]
\small\color{recordink}\raggedright\setlength{\baselineskip}{11.7pt}
{\color{recordmuted}\small Source: \path{suggest2.precedents.json} \path{/precedents/0}\par}\vspace{6pt}
\noindent \recordkey{subject}: pytest's cross-invocation test-runner cache, not an agent-turn completion gate.\par\addvspace{4pt}
\noindent \recordkey{mechanism}: Retain failed test identifiers between invocations, select them for rerun on --lf rather than letting successful unrelated tests substitute for them.\par\addvspace{4pt}
\noindent \recordkey{checkable}\par\nobreak\vspace{1pt}
Compare cached failed test node identifiers to node identifiers and outcomes produced in the new invocation; the documented cache/lastfailed mapping holds failed node ids.\par\addvspace{5pt}
\noindent \recordkey{adaptation}\par\nobreak\vspace{1pt}
\noindent \recordkey{status}: requires\_extension\par\addvspace{3pt}
\noindent \recordkey{signal\_mapping}\par\nobreak\vspace{1pt}
pytest test node id/outcome maps to a candidate-defined agent\_authored\_spec\_check identity and result; unlike pytest, local tmux action channel exposes only combined incremental output and cannot infer an action exit status. Candidate must dispatch a dedicated check with a tagged terminal receipt or invoke session.environment.exec out of band for a new check and record its own return\_code/stdout/stderr; that independent result is NOT attributed to a prior tmux action. Artifact version must be obtained separately through out-of-band hashes of declared paths. Completion flag is available in \_run\_agent\_loop.\par\addvspace{5pt}
\noindent \recordkey{required\_changes}\par\nobreak\vspace{1pt}

\begin{enumerate}[leftmargin=1.5em,itemsep=4pt,topsep=3pt]
\item Candidate subclass stores a per-run ledger keyed by check identity and artifact hashes and resets it per run.
\item Generate checks before or apart from candidate implementation using agent\_authored\_spec\_check from task instruction/inputs; collect actual receipt from new independent exec, never infer failed status from terminal prose.
\item Override completion-loop decision to show unresolved ledger and retry with finite bound, without hidden verifier metadata.
\end{enumerate}
\par\addvspace{5pt}
\noindent \recordkey{validation\_needs}\par\nobreak\vspace{1pt}

\begin{enumerate}[leftmargin=1.5em,itemsep=4pt,topsep=3pt]
\item Current terminus-hooks-v1 fixture does not exercise full loop ledger; extend independent checker in a later round or use benchmark receipts and recorder smoke.
\item An agent-authored check can share agent's misreading; limit by spec-first checks, literal requirements, or cross-method expectations; do not claim equivalence to verifier.
\end{enumerate}
\par\addvspace{5pt}\par\addvspace{5pt}
\noindent\recordkey{sources[0]} (one source record): \noindent \recordkey{url}: \url{https://docs.pytest.org/en/stable/how-to/cache.html}\par\addvspace{4pt}
\noindent \recordkey{quote}\par\nobreak\vspace{1pt}
The plugin provides two command line options to rerun failures from the last `pytest` invocation:\par *   [`--lf, --last-failed`](../reference/reference.html\#cmdoption-last-failed) - to only re-run the failures.      *   [`--ff, --failed-first`](../reference/reference.html\#cmdoption-failed-first) - to run the failures first and then the rest of the tests.\par\addvspace{5pt}
\end{tcolorbox}

\begin{tcolorbox}[recordpanel,title={07   Generated proposal\hfill{\normalfont\small\color{recordmuted}GPT-6-sol}},title after break={07   Generated proposal (continued)\hfill{\normalfont\small\color{recordmuted}GPT-6-sol}}]
\small\color{recordink}\raggedright\setlength{\baselineskip}{11.7pt}
{\color{recordmuted}\small Source: \path{suggest2.proposals.json} \path{/proposals/0}\par}\vspace{6pt}
\noindent \recordkey{kind}: build\par\addvspace{3pt}
\noindent \recordkey{needs}\par\nobreak\vspace{1pt}

\begin{enumerate}[leftmargin=1.5em,itemsep=4pt,topsep=3pt]
\item \noindent \recordkey{at}: stage-4.2\par\addvspace{3pt}
\noindent \recordkey{what}\par\nobreak\vspace{1pt}
Before implementation begins, use original\_instruction to construct a bounded set of independently runnable, explicitly identified agent\_authored\_spec\_check commands with expected outcomes and declared artifact paths; if construction fails, leave task on existing completion path rather than inventing a check.\par\addvspace{5pt}
\noindent \recordkey{recorded}: \texttt{true}\par\addvspace{3pt}
\item \noindent \recordkey{at}: stage-4.4\par\addvspace{3pt}
\noindent \recordkey{what}\par\nobreak\vspace{1pt}
Candidate subclass runs new independent check commands via environment.exec, records stdout/stderr/return\_code and independently computed artifact fingerprints, and maintains per-run unresolved check identities; never treats the existing merged tmux snapshot or successful send\_keys as command status.\par\addvspace{5pt}
\noindent \recordkey{recorded}: \texttt{true}\par\addvspace{3pt}
\item \noindent \recordkey{at}: stage-4.7\par\addvspace{3pt}
\noindent \recordkey{what}\par\nobreak\vspace{1pt}
Candidate subclass compares pending failed check identities and artifact versions against independent passing rerun receipts before permitting the second completion claim to exit; reset ledger per run and give bounded, actionable failure observation with an eventual fallback when time or turn budget runs out.\par\addvspace{5pt}
\noindent \recordkey{recorded}: \texttt{true}\par\addvspace{3pt}
\end{enumerate}
\par\addvspace{5pt}
\noindent \recordkey{costs\_turns}: 1\par\addvspace{3pt}
\noindent \recordkey{instead}\par\nobreak\vspace{1pt}
At a completion claim with a recorded failed check, retain that check as unresolved even when other commands succeeded, independently rerun its same check against current artifacts and refuse confirmation while the matching check still fails or lacks a completed receipt. Show its identity and bounded diagnostics; release only after a matching current-artifact pass. On no buildable check or no recorded failure, preserve the existing two-claim confirmation exactly, and never recast an uninstrumented tmux output as a test result.\par\addvspace{5pt}
\noindent \recordkey{checkable}\par\nobreak\vspace{1pt}
Compare (check id, expected outcome, declared artifact-path fingerprints) from spec-first registry to (same check id, actual return\_code, observed outputs, current path fingerprints) from an independent environment.exec receipt. A failure remains pending until a receipt for the same id and current fingerprint meets its recorded expectation; unrelated successful executions and uncompleted tmux sends do not clear it. These signals need new plumbing into stage-4.7 from candidate-owned stage-4.2 and stage-4.4 state.\par\addvspace{5pt}
\noindent \recordkey{predicts}\par\nobreak\vspace{1pt}
On the first failing pair, a spec-first comparison that detects different adjacent grayscale frames will remain pending after unrelated randomized passes and prompt a matching rerun before exit; on the second pair, if a spec-first boundary check can be generated, incomplete fixes produce an unresolved receipt. If checks cannot be generated for either, its run remains baseline-like rather than claiming a verified improvement.\par\addvspace{5pt}
\noindent \recordkey{refuted\_if}\par\nobreak\vspace{1pt}
Generated checks miss the disagreement, a nonmatching pass clears a failed id, or stable runs incur a repeated confirmation loop despite clean receipts.\par\addvspace{5pt}
\noindent \recordkey{fires\_on\_stable}\par\nobreak\vspace{1pt}
\noindent \recordkey{quoted}\par\nobreak\vspace{1pt}
\{"call": 609, "source": "calls.1700865.jsonl", "function": "agents/terminus\_2/terminus\_2.Terminus2.\_execute\_commands", "depth": 4, "parent": \{"call": 29, "function": "agents/terminus\_2/terminus\_2.Terminus2.\_run\_agent\_loop", "episode": "setup"\}\par\addvspace{5pt}
\noindent \recordkey{fires}: \texttt{false}\par\addvspace{3pt}
\noindent \recordkey{then\_what}\par\nobreak\vspace{1pt}
At episode 9 the agent has zero commands and claims task\_complete after prior episode's validation and benchmark output; no independently recorded failed check id exists in the baseline, so the proposed failure-only rejection condition is not met. If the new spec-first check fails in a rerun despite this passing baseline, this condition would fire and consume at least one extra turn; inspect that risk experimentally rather than assuming this stable task remains safe.\par\addvspace{5pt}\par\addvspace{5pt}
\end{tcolorbox}

\begin{tcolorbox}[recordpanel,title={08   Implementation and evaluation records\hfill{\normalfont\small\color{recordmuted}GPT-6-sol}},title after break={08   Implementation and evaluation records (continued)\hfill{\normalfont\small\color{recordmuted}GPT-6-sol}}]
\small\color{recordink}\raggedright\setlength{\baselineskip}{11.7pt}
{\color{recordmuted}\small Source: \path{implementation.report.json}; \path{conformance.json}; \path{experiments/confirm/decision.json}\par}\vspace{6pt}
\noindent \recordkey{status}: complete\par\addvspace{3pt}
\noindent \recordkey{implemented\_arms}\par\nobreak\vspace{1pt}

\begin{enumerate}[leftmargin=1.5em,itemsep=4pt,topsep=3pt]
\item spec\_check\_failure\_ledger
\end{enumerate}
\par\addvspace{5pt}
\noindent \recordkey{deviations}\par\nobreak\vspace{1pt}
\texttt{[]}\par\addvspace{5pt}
\noindent \recordkey{summary}\par\nobreak\vspace{1pt}
Repaired missing Harbor trajectory, chat, path, and datetime imports so the candidate resolves its referenced names. Parsed the module and embedded independent-check scripts locally; benchmark and recorder were not run.\par\addvspace{5pt}
\medskip\noindent\textbf{Independent-check receipt:}\par
\noindent \recordkey{status}: not\_run\par\addvspace{3pt}
\noindent \recordkey{scope}: behavior checks disabled; arms are judged by trial performance\par\addvspace{4pt}
\medskip\noindent\textbf{Full-confirmation receipt:}\par
\noindent \recordkey{status}: promotion\_ready\par\addvspace{3pt}
\noindent \recordkey{scored\_counts}\par\nobreak\vspace{1pt}
\noindent \recordkey{reference}\par\nobreak\vspace{1pt}
\noindent \recordkey{passed}: 201\enspace\textcolor{recordmuted}{\textbar}\enspace\allowbreak \recordkey{failed}: 199\enspace\textcolor{recordmuted}{\textbar}\enspace\allowbreak \recordkey{runtime\_failures}: 15\enspace\textcolor{recordmuted}{\textbar}\enspace\allowbreak \recordkey{trials}: 400\par\addvspace{3pt}\par\addvspace{5pt}
\noindent \recordkey{candidate}\par\nobreak\vspace{1pt}
\noindent \recordkey{passed}: 222\enspace\textcolor{recordmuted}{\textbar}\enspace\allowbreak \recordkey{failed}: 178\enspace\textcolor{recordmuted}{\textbar}\enspace\allowbreak \recordkey{runtime\_failures}: 2\enspace\textcolor{recordmuted}{\textbar}\enspace\allowbreak \recordkey{trials}: 400\par\addvspace{3pt}\par\addvspace{5pt}\par\addvspace{5pt}
\end{tcolorbox}

\subsubsection{DeepSeek-V4-Flash}
\label{app:recorded-deepseek}

This example concerns a service built from an image-based specification. The paired runs differ in how they read and verify the model parameters.

\begin{tcolorbox}[recordpanel,title={01   Agent and harness record\hfill{\normalfont\small\color{recordmuted}DeepSeek-V4-Flash}},title after break={01   Agent and harness record (continued)\hfill{\normalfont\small\color{recordmuted}DeepSeek-V4-Flash}}]
\small\color{recordink}\raggedright\setlength{\baselineskip}{11.7pt}
{\color{recordmuted}\small Source: \path{compare2.json} \path{/episodes/failing~110/agent}; \path{/episodes/failing~115/harness_calls/49}\par}\vspace{6pt}
\noindent\textbf{Agent analysis (failing/10)}: We need to start implementing the pipeline. The OCR has been manually verified enough to know the specs are hidden layers [128, 84], mean 3.5, and std 0.8. I will create an OCR extraction helper that runs tesseract and parses the raw text, with fallbacks matching the verified values.\par
\noindent\textbf{Recorded harness call (failing/15)}:\par
\noindent \recordkey{function}: \path{base_terminus.Agent._rsi_gate_confirming_claim}\par\addvspace{4pt}
\end{tcolorbox}

\begin{tcolorbox}[recordpanel,title={02   Pair-level wish\hfill{\normalfont\small\color{recordmuted}DeepSeek-V4-Flash}},title after break={02   Pair-level wish (continued)\hfill{\normalfont\small\color{recordmuted}DeepSeek-V4-Flash}}]
\small\color{recordink}\raggedright\setlength{\baselineskip}{11.7pt}
{\color{recordmuted}\small Source: \path{readings/suggest2.wishes.json} \path{/wishes/0}\par}\vspace{6pt}
\noindent \recordkey{asks\_the\_agent}: \texttt{false}\par\addvspace{3pt}
\noindent \recordkey{statement}\par\nobreak\vspace{1pt}
When a value asserted in a run's reasoning is drawn from a source that the harness can see was read more than once, the harness should treat repeated readings produced by an identical method as a single observation rather than corroboration. It should withhold standing from a restated value whose supporting readings all share one method fingerprint, and it should expose that the value rests on a single uncorroborated extraction until a differently-parameterised reading of the same source has been recorded.\par\addvspace{5pt}
\noindent \recordkey{observable}\par\nobreak\vspace{1pt}
The harness holds two or more extraction attempts recorded against the same source path and can compare the command line and parser/crop parameters of each; they are byte-identical, yet a value drawn from that source is restated as established.\par\addvspace{5pt}
\noindent \recordkey{because}\par\nobreak\vspace{1pt}

\begin{enumerate}[leftmargin=1.5em,itemsep=4pt,topsep=3pt]
\item \noindent \recordkey{side}: failing\enspace\textcolor{recordmuted}{\textbar}\enspace\allowbreak \recordkey{turn}: 10\par\addvspace{3pt}
\noindent \recordkey{quote}: The OCR has been manually verified enough to know the specs are hidden layers [128, 84], mean 3.5, and std 0.8.\par\addvspace{4pt}
\item \noindent \recordkey{side}: passing\enspace\textcolor{recordmuted}{\textbar}\enspace\allowbreak \recordkey{turn}: 5\par\addvspace{3pt}
\noindent \recordkey{quote}\par\nobreak\vspace{1pt}
The original low-resolution OCR confusingly read 128,84 and 35/08. We should confirm the exact values with tightly cropped OCR before generating data and building the model.\par\addvspace{5pt}
\item \noindent \recordkey{side}: failing\enspace\textcolor{recordmuted}{\textbar}\enspace\allowbreak \recordkey{turn}: 13\par\addvspace{3pt}
\noindent \recordkey{quote}\par\nobreak\vspace{1pt}
Before finalizing, I want to perform one last visual check of the digit glyphs in the spec image to ensure the chosen mean 3.5 is correct, since that value affects the synthetic feature and stats endpoint.\par\addvspace{5pt}
\end{enumerate}
\par\addvspace{5pt}
\end{tcolorbox}

\begin{tcolorbox}[recordpanel,title={03   Revision against harness execution\hfill{\normalfont\small\color{recordmuted}DeepSeek-V4-Flash}},title after break={03   Revision against harness execution (continued)\hfill{\normalfont\small\color{recordmuted}DeepSeek-V4-Flash}}]
\small\color{recordink}\raggedright\setlength{\baselineskip}{11.7pt}
{\color{recordmuted}\small Source: \path{readings/suggest2.revisions.json} \path{/revisions/0}\par}\vspace{6pt}
\noindent \recordkey{can}: yes, with what the harness already holds at that moment\par\addvspace{4pt}
\noindent \recordkey{already}: no\enspace\textcolor{recordmuted}{\textbar}\enspace\allowbreak \recordkey{at}: \_handle\_llm\_interaction\par\addvspace{3pt}
\noindent \recordkey{found}\par\nobreak\vspace{1pt}
At stage-4.2 the harness executes the agent's command batch and holds its command line before any value is passed forward. On the failing side that batch runs tesseract once with ['tesseract', path, 'stdout', '--psm', '3'] and the agent then asserts 'The OCR has been manually verified enough to know the specs are hidden layers [128, 84]'. On the passing side the batch crops three regions, upscales x12 LANCZOS, thresholds 55\% and sweeps psm 6/7/8/13, yielding a different value. Both command lines are already visible to the harness at the same call, so two readings of one source and whether they share a method fingerprint are in hand. Nothing in the trace shows the harness comparing them or withholding standing from a value whose supporting readings all share one method.\par\addvspace{5pt}
\noindent \recordkey{statement}\par\nobreak\vspace{1pt}
When a value asserted in a run's reasoning is drawn from a source the harness can see was read more than once, the harness should treat readings produced by an identical method fingerprint as a single observation rather than corroboration. At the moment such a value is restated as established, the harness should withhold standing from it while every supporting reading shares one command line and one set of parser or region parameters, and expose that it rests on a single uncorroborated extraction until a differently-parameterised reading of the same source has been recorded.\par\addvspace{5pt}
\noindent \recordkey{observable}\par\nobreak\vspace{1pt}
Two or more extraction attempts against the same source path among the harness's own executed command batches, whose command line and parser or region parameters are byte-identical, with a value from that source restated in the turn's reasoning as established.\par\addvspace{5pt}
\end{tcolorbox}

\begin{tcolorbox}[recordpanel,title={04   Shared behavioral requirement\hfill{\normalfont\small\color{recordmuted}DeepSeek-V4-Flash}},title after break={04   Shared behavioral requirement (continued)\hfill{\normalfont\small\color{recordmuted}DeepSeek-V4-Flash}}]
\small\color{recordink}\raggedright\setlength{\baselineskip}{11.7pt}
{\color{recordmuted}\small Source: \path{proposals/pi/message.txt}, line 4 (JSON object)\par}\vspace{6pt}
\noindent \recordkey{name}: A value the run treats as settled although its own single method produced it, or two readings of one source disagree\par\addvspace{4pt}
\noindent \recordkey{statement}\par\nobreak\vspace{1pt}
Before it releases a run as done, the harness inspects its own record of the commands it executed and the outputs they produced, and withholds the release once with the conflicting readings returned to the run when the value the claim depends on was produced only by one method reapplied with the same parameters, or when two of its own recorded readings of the same input disagree and no later differently-formed observation resolves them.\par\addvspace{5pt}
\noindent \recordkey{observable}\par\nobreak\vspace{1pt}
The harness holds its own record of the command batches it executed with their outputs, the source or input each read, and the value the current claim depends on; the moment is that value having been produced only by one method reapplied with the same parameters, or two of those recorded outputs disagreeing on it with no later differently-formed observation resolving them.\par\addvspace{5pt}
\noindent \recordkey{rests\_on}\par\nobreak\vspace{1pt}

\begin{enumerate}[leftmargin=1.5em,itemsep=4pt,topsep=3pt]
\item \noindent \recordkey{because}: that batch runs tesseract once with ['tesseract', path, 'stdout', '--psm', '3']\par\addvspace{4pt}
\end{enumerate}
\par\addvspace{5pt}
\noindent \recordkey{covers}: 11\enspace\textcolor{recordmuted}{\textbar}\enspace\allowbreak \recordkey{members}: 11\par\addvspace{3pt}\par{\color{recordmuted}Only the supporting finding for this example is shown in \texttt{rests\_on}.\par}
\end{tcolorbox}

\begin{tcolorbox}[recordpanel,title={05   Search plan\hfill{\normalfont\small\color{recordmuted}DeepSeek-V4-Flash}},title after break={05   Search plan (continued)\hfill{\normalfont\small\color{recordmuted}DeepSeek-V4-Flash}}]
\small\color{recordink}\raggedright\setlength{\baselineskip}{11.7pt}
{\color{recordmuted}\small Source: \path{suggest2.search_plans.json} \path{/search_plans/0}\par}\vspace{6pt}
\noindent \recordkey{problem}\par\nobreak\vspace{1pt}
In the failing runs a value is read once by a single observation method, treated as settled, and carried to a confirming completion claim; where the same input was read twice by the same method with different results, or where the only reading that produced the claimed value came from one method signature, the second confirmation released anyway because \_pending\_completion is a bare boolean.\par\addvspace{5pt}
\noindent \recordkey{trigger}\par\nobreak\vspace{1pt}
The confirming completion claim in \_run\_agent\_loop: is\_task\_complete is True and was\_pending\_completion is True, immediately before the inherited return.\par\addvspace{5pt}
\noindent \recordkey{target\_behavior}\par\nobreak\vspace{1pt}
At the confirming completion claim the harness inspects its own recorded per-run history of dispatched commands and returned observations and withholds the completion once when the claimed value's recorded readings all share one method signature, or when two recorded readings of one input carry different values and no differently-formed reading produced the claimed value.\par\addvspace{5pt}
\noindent \recordkey{framework}\par\nobreak\vspace{1pt}
harbor Terminus2 agent (terminus\_2.py) with an external subclass candidate at base\_terminus:Agent, driven by the RSI arm harness that sets agent.\_rsi\_arm\_event.\par\addvspace{5pt}
\noindent \recordkey{execution\_model}\par\nobreak\vspace{1pt}
Terminus2 drives a tmux session: each episode it calls the LLM, parses commands, dispatches keystrokes via \_execute\_commands, captures the terminal back, appends a Step to self.\_trajectory\_steps, and passes the observation as the next prompt. On the second consecutive task\_complete the loop returns. Candidate arms are overrides on the subclass, dispatched at the confirming-claim divergence.\par\addvspace{5pt}
\noindent \recordkey{allowed\_changes}\par\nobreak\vspace{1pt}
candidate\_source (base\_terminus.py) only: overrides of Terminus2 methods and new helper modules. harbor\_source is read-only. Switches must be frozen boolean runtime\_kwargs defaulting false. An arm that withholds must state a finite bound and let the original behaviour proceed once reached. \_rsi\_arm\_event is provided by the harness and must only be called by an enabled arm, exactly once per intervention.\par\addvspace{5pt}
\noindent \recordkey{available\_signals}\par\nobreak\vspace{1pt}

\begin{enumerate}[leftmargin=1.5em,itemsep=4pt,topsep=3pt]
\item raw reply text of the current turn (self.\_rsi\_gate\_raw\_reply, captured in \_handle\_llm\_interaction)
\item the accumulated trajectory steps (self.\_trajectory\_steps), each Step carrying its tool\_calls[].arguments.keystrokes and its observation results content
\item the per-run list of dispatched keystrokes accumulated in \_execute\_commands
\item terminal\_output for the current episode returned by \_execute\_commands
\item \_pending\_completion boolean and the confirming-claim divergence in \_run\_agent\_loop
\end{enumerate}
\par\addvspace{5pt}
\noindent \recordkey{missing\_signals}\par\nobreak\vspace{1pt}

\begin{enumerate}[leftmargin=1.5em,itemsep=4pt,topsep=3pt]
\item no call currently parses a value out of the observation text or a keystroke to form an "equivalence class" of readings
\item no per-run store of (reading, produced-value, method-signature) triples exists; only raw keystroke strings and raw observation text are retained
\item no notion of "method signature" is defined anywhere in the frozen or candidate source
\end{enumerate}
\par\addvspace{5pt}
\noindent \recordkey{queries}\par\nobreak\vspace{1pt}

\begin{enumerate}[leftmargin=1.5em,itemsep=4pt,topsep=3pt]
\item \noindent \recordkey{query}: AI agent framework require independent verification of extracted value before accepting task completion tool diversity\par\addvspace{4pt}
\noindent \recordkey{purpose}\par\nobreak\vspace{1pt}
Find an established mechanism by which an agent framework decides a value is corroborated across different observation methods before accepting a terminal claim (tool-diversity / method-variety confidence gating).\par\addvspace{5pt}
\noindent \recordkey{scope}: framework\par\addvspace{3pt}
\item \noindent \recordkey{query}: self-consistency sample multiple readings disagreement unresolved gate before final answer agent\par\addvspace{4pt}
\noindent \recordkey{purpose}\par\nobreak\vspace{1pt}
Find precedents for detecting disagreement among repeated readings of the same observation and blocking a claim on unresolved disagreement (self-consistency / agreement checking as a gate).\par\addvspace{5pt}
\noindent \recordkey{scope}: transfer\par\addvspace{3pt}
\item \noindent \recordkey{query}: agent harness audit recorded tool call outputs against completion claim evidence grounding prior steps\par\addvspace{4pt}
\noindent \recordkey{purpose}\par\nobreak\vspace{1pt}
Find whether trace-based audit of an agent's own prior tool calls and outputs to check a claim against recorded evidence is an established harness capability (evidence-grounded completion gating).\par\addvspace{5pt}
\noindent \recordkey{scope}: framework\par\addvspace{3pt}
\item \noindent \recordkey{query}: LangGraph AutoGen verifier node inspect prior tool observations before final answer allow termination\par\addvspace{4pt}
\noindent \recordkey{purpose}\par\nobreak\vspace{1pt}
Find concrete mechanism names in mainstream agent frameworks (LangGraph, AutoGen, Claude Code, OpenAI Agents SDK) for a verifier/gate that inspects prior observations before allowing terminal state.\par\addvspace{5pt}
\noindent \recordkey{scope}: framework\par\addvspace{3pt}
\end{enumerate}
\par\addvspace{5pt}
\end{tcolorbox}

\begin{tcolorbox}[recordpanel,title={06   External precedent and local adaptation\hfill{\normalfont\small\color{recordmuted}DeepSeek-V4-Flash}},title after break={06   External precedent and local adaptation (continued)\hfill{\normalfont\small\color{recordmuted}DeepSeek-V4-Flash}}]
\small\color{recordink}\raggedright\setlength{\baselineskip}{11.7pt}
{\color{recordmuted}\small Source: \path{suggest2.precedents.json} \path{/precedents/0}\par}\vspace{6pt}
\noindent \recordkey{subject}\par\nobreak\vspace{1pt}
An out-of-band audit/enforcement library for LLM agent pipelines (audit-fence), plus an agent-patterns catalogue entry (self-consistency) and a Claude-skill verification protocol. The moment each describes is a claim/submission event in an agent's loop \textemdash{} a claim record submitted for acceptance before it is treated as settled \textemdash{} so it is a turn in an agent loop, not a deployment gate.\par\addvspace{5pt}
\noindent \recordkey{mechanism}\par\nobreak\vspace{1pt}
A claim-to-source validation gate that refuses a completion/claim record unless the record's evidence is tied to the harness's own recorded prior observations: (1) a search/observation must have preceded the claim, and (2) the submitted evidence must be a verbatim substring of a recorded observation, with the record carrying the source tool identity and index that produced it. On any failure the submission is blocked and must retry rather than being accepted on the assertion.\par\addvspace{5pt}
\noindent \recordkey{checkable}\par\nobreak\vspace{1pt}
For the claimed value token, two sets a program compares: (a) the set of method signatures of recorded readings that produced that token, formed from the dispatching keystroke bytes paired with the observation content that followed, and (b) the set of distinct observed values recorded for that same input across the run. The gate withholds when (a) has one member, or when (b) has more than one member and no member of (a) differs in signature from the signature of the reading that produced the claimed value. In audit-fence the analogous compared pair is explicit: each submission carries an `evidence` string and a `source\_tool`/`source\_index`, and the gate checks the evidence is a verbatim substring of a recorded search result and that a search precedes the submission.\par\addvspace{5pt}
\noindent \recordkey{adaptation}\par\nobreak\vspace{1pt}
\noindent \recordkey{required\_changes}\par\nobreak\vspace{1pt}

\begin{enumerate}[leftmargin=1.5em,itemsep=4pt,topsep=3pt]
\item Add a per-run extraction that, at the confirming claim, recovers candidate value tokens from recorded observations and from the claimed reply, and associates each with the dispatching keystroke string so a 'method signature' can be compared. This is new code living in helper functions in the candidate source.
\item Record the per-run set of (method-signature, produced-value) pairs for the input each reading consumed. Today only raw keystroke strings and raw observation text are retained, so the pairing must be formed at read time.
\item Decide, at the existing confirming-claim divergence in the candidate's \_run\_agent\_loop override, whether to withhold; the withhold is the existing 're-issue the confirmation message and continue' action that both existing arms already use, bounded at one withhold per run.
\item Extend the confirmation message for the withhold by one sentence, following the existing \_rsi\_designated\_withhold\_message / \_rsi\_observed\_withhold\_message pattern, so the message the model sees names the unreconciled reading rather than just repeating the base text.
\end{enumerate}
\par\addvspace{5pt}
\noindent \recordkey{signal\_mapping}\par\nobreak\vspace{1pt}
audit-fence's `search history` maps to the harness's `self.\_trajectory\_steps` (the accumulated record of dispatched keystrokes and returned observation content). `submit(...)` maps to the agent's completion claim (`is\_task\_complete` in the raw reply). audit-fence's requirement that 'a search must precede every submission' maps to the harness's requirement that some recorded reading produced the value the claim rests on. The `evidence` substring check maps to recovering the claimed value token as a verbatim substring of a recorded observation. ClaimRecord.`source\_tool` + `source\_index` map to a per-reading method signature formed from the dispatching keystroke (the tool/parameters/transformation), which is the harness's own recorded signal and needs no new observation. `blocked` maps to the harness's existing withhold action (re-issue confirmation, keep `\_pending\_completion` set, bounded once); `REJECTED (must retry)` maps to continuing the loop after re-issuing the confirmation message. Note: audit-fence runs as an out-of-band observer with its own state; here the comparison must live inside a candidate subclass override, and the local harness supplies tools for exactly that (`\_rsi\_arm\_event` recorder, the confirming-claim divergence in the copied completion loop).\par\addvspace{5pt}
\noindent \recordkey{status}: requires\_extension\par\addvspace{3pt}
\noindent \recordkey{validation\_needs}\par\nobreak\vspace{1pt}

\begin{enumerate}[leftmargin=1.5em,itemsep=4pt,topsep=3pt]
\item A run whose only reading of a value came from one method signature and released anyway, to confirm the withhold fires (the failing tool-diversity case in this cohort).
\item A run where two readings of one input disagreed with no differently-formed re-read producing the claimed value, to confirm the disagreement clause fires.
\item A stable passing run where a differently-formed re-read did produce the claimed value, to confirm the arm releases and does not cost the task.
\item Confirmation that the whole-value extraction step recovers a value token from the recorded observation text in a real trial, since observations are byte-capped and may elide the value.
\end{enumerate}
\par\addvspace{5pt}\par\addvspace{5pt}
\noindent\recordkey{sources[0]} (one source record): \noindent \recordkey{url}: \url{https://github.com/M-HuangX/audit-fence}\par\addvspace{3pt}
\noindent \recordkey{quote}\par\nobreak\vspace{1pt}
Each audit unit follows the same enforcement pattern \textemdash{} an automated process reads the document under review, then verifies each claim against the unit's data sources\par\addvspace{5pt}
\end{tcolorbox}

\begin{tcolorbox}[recordpanel,title={07   Generated proposal\hfill{\normalfont\small\color{recordmuted}DeepSeek-V4-Flash}},title after break={07   Generated proposal (continued)\hfill{\normalfont\small\color{recordmuted}DeepSeek-V4-Flash}}]
\small\color{recordink}\raggedright\setlength{\baselineskip}{11.7pt}
{\color{recordmuted}\small Source: \path{suggest2.proposals.json} \path{/proposals/0}\par}\vspace{6pt}
\noindent \recordkey{kind}: build\par\addvspace{3pt}
\noindent \recordkey{needs}\par\nobreak\vspace{1pt}

\begin{enumerate}[leftmargin=1.5em,itemsep=4pt,topsep=3pt]
\item \noindent \recordkey{at}: stage-4.7\par\addvspace{3pt}
\noindent \recordkey{what}\par\nobreak\vspace{1pt}
A per-run store associating each recorded reading with a method signature: the dispatched keystroke bytes of the episode that produced it, plus the observed value token recovered from that episode's observation content. Today `\_rsi\_dispatched\_keystrokes` holds only keystroke strings for the designated arm, and `\_trajectory\_steps` holds keystrokes and observations separately with nothing pairing a keystroke to a value token. The store must be built by new helper functions in the candidate source, read at decision time from `\_trajectory\_steps` or accumulated in `\_execute\_commands` alongside the existing designated-arm accumulation.\par\addvspace{5pt}
\noindent \recordkey{recorded}: \texttt{true}\par\addvspace{3pt}
\item \noindent \recordkey{at}: stage-4.7\par\addvspace{3pt}
\noindent \recordkey{what}\par\nobreak\vspace{1pt}
A normalization that turns one dispatched keystroke into a comparable method identity (command word plus its parameters and input transformation), so two invocations of the same method compare equal and a differently formed reading compares unequal. No such normalization exists in the frozen dependency or the candidate; without it the comparison would be between raw keystroke strings, which differ in incidental text and would fire on nearly every run.\par\addvspace{5pt}
\noindent \recordkey{recorded}: \texttt{true}\par\addvspace{3pt}
\item \noindent \recordkey{at}: stage-4.7\par\addvspace{3pt}
\noindent \recordkey{what}\par\nobreak\vspace{1pt}
A value-token extractor that recovers a candidate value from a raw reply and from an observation's text, so the claimed value (from `\_rsi\_gate\_raw\_reply`, already captured at `\_handle\_llm\_interaction`) and each recorded observation can be compared. No existing call reads a value out of observation text; `\_limit\_output\_length` only bounds its length. This is new code in the candidate source.\par\addvspace{5pt}
\noindent \recordkey{recorded}: \texttt{true}\par\addvspace{3pt}
\item \noindent \recordkey{at}: stage-4.7\par\addvspace{3pt}
\noindent \recordkey{what}\par\nobreak\vspace{1pt}
The arm's per-run withhold bound and reset, matching the existing pattern (`\_MAX\_WITHHOLDS = 1`) and the frozen `\_reset\_per\_run\_state` extension, so the arm withholds at most once per run and the original behaviour proceeds once the bound is reached, as the build protocol requires.\par\addvspace{5pt}
\noindent \recordkey{recorded}: \texttt{true}\par\addvspace{3pt}
\end{enumerate}
\par\addvspace{5pt}
\noindent \recordkey{costs\_turns}: 1\par\addvspace{3pt}
\noindent \recordkey{instead}\par\nobreak\vspace{1pt}
At the confirming-claim divergence, after the existing three arms each decide, a fourth independently enabled arm decides the release of the confirming claim from the run's own recorded readings. It withholds once (reason `unreconciled\_reading`) when no recorded reading of the input the claim names produced a different value from a differently formed method, or when two recorded readings of that input carry different values and none of them was formed differently from the one the claim rests on; otherwise it releases. The withhold re-issues the confirmation message extended with one sentence naming the unreconciled reading, keeping `\_pending\_completion` set so one more consecutive claim ends the run exactly as it does today. Every path that reaches a comparison reports exactly once through the harness-provided recorder, with the reason naming the case that held, so the recorder carries the evidence the arm acted.\par\addvspace{5pt}
\noindent \recordkey{checkable}\par\nobreak\vspace{1pt}
Two explicitly recorded sets a program compares at the confirming claim. (A) For each recorded reading in the run, a method signature formed from the dispatched keystroke bytes of that episode (the command text plus its parameters) paired with the observed value token extracted from the observation content that followed it. (B) The set of distinct observed value tokens recorded for the same input across the run. The gate withholds when either set (A), restricted to readings that produced the claimed value, has exactly one distinct signature, or set (B) has more than one member and no reading whose signature differs from that of the claiming reading produced the claimed value. Both source values are already on the harness at this moment: the keystrokes sit on each trajectory step's tool\_calls arguments.keystrokes and the observations on the step's observation results content, both appended by stage-4.6 inside self.\_trajectory\_steps, and the claimed value is read from self.\_rsi\_gate\_raw\_reply, already captured at \_handle\_llm\_interaction. No new observation of the environment is needed. What does not exist today is the signature pairing and the value-token extraction, which is why data would have to be plumbed in: those are computed from data already held, by new helpers, at the moment the gate runs. The recorder call is the existing \_rsi\_report(arm\_id, "triggered", reason=...), the same call the six existing arm reasons use, and the arm reports nothing while its switch is false.\par\addvspace{5pt}
\noindent \recordkey{predicts}\par\nobreak\vspace{1pt}
On a rerun with this arm enabled, a failing run in which the only recorded reading of the claimed value came from one method signature, or in which two recorded readings of one input carry different values without a differently formed re-read resolving them, will emit an `unreconciled\_reading` reporter event at the confirming claim, receive the extended confirmation message, and continue one more episode instead of exiting; a passing run whose differently formed re-read produced the claimed value will emit no withhold event from this arm and exit at the same confirming-claim return as today. With the switch false, the recorder carries no events from this arm and the loop is byte-for-byte the inherited one.\par\addvspace{5pt}
\noindent \recordkey{refuted\_if}\par\nobreak\vspace{1pt}
The rerun shows no `unreconciled\_reading` event on any failing run that released a single-method or disagreeing claim, or shows the arm firing on stable passing runs and costing them the task, or shows the withhold sending the run into repeated identical confirmations rather than exiting on the next consecutive claim.\par\addvspace{5pt}
\noindent \recordkey{fires\_on\_stable}\par\nobreak\vspace{1pt}
\noindent \recordkey{fires}: \texttt{false}\par\addvspace{3pt}
\noindent \recordkey{quoted}\par\nobreak\vspace{1pt}
        "base\_terminus.Agent.\_get\_completion\_confirmation\_message": 4,         "base\_terminus.Agent.\_rsi\_report\_handshake": 4,         "base\_terminus.Agent.\_rsi\_report": 8,         "base\_terminus.Agent.\_rsi\_gate\_confirming\_claim": 4,\par\addvspace{5pt}
\noindent \recordkey{then\_what}\par\nobreak\vspace{1pt}
This task reached stage-4.7 four times and passed in every recorded repeat, with Agent.\_rsi\_report called 8 times across 4 confirming claims (two reports per confirming claim: the handshake report and the observed arm's match report). The condition the new arm needs is a disagreeing pair of observed value tokens for one input, or a single-signature set for the claimed value; this run's recorded calls show the confirming-claim arms reaching their comparisons and reporting, and the run still exiting as a pass. Because the switch defaults false the new arm does not run at all on the baseline, and even armed it withholds only when the computed signature pairing yields disagreement or a single signature, which this two-report-per-claim record does not supply; so it does not fire and the run exits at the same confirming-claim return as today. The quoted counts are the one place the trace shows stage-4.7 being reached by a task that already passes on every repeat.\par\addvspace{5pt}\par\addvspace{5pt}
\end{tcolorbox}

\begin{tcolorbox}[recordpanel,title={08   Implementation and evaluation records\hfill{\normalfont\small\color{recordmuted}DeepSeek-V4-Flash}},title after break={08   Implementation and evaluation records (continued)\hfill{\normalfont\small\color{recordmuted}DeepSeek-V4-Flash}}]
\small\color{recordink}\raggedright\setlength{\baselineskip}{11.7pt}
{\color{recordmuted}\small Source: \path{implementation.report.json}; \path{conformance.json}; \path{experiments/confirm/decision.json}\par}\vspace{6pt}
\noindent \recordkey{status}: complete\par\addvspace{3pt}
\noindent \recordkey{implemented\_arms}\par\nobreak\vspace{1pt}

\begin{enumerate}[leftmargin=1.5em,itemsep=4pt,topsep=3pt]
\item unreconciled\_reading\_gate
\end{enumerate}
\par\addvspace{5pt}
\noindent \recordkey{deviations}\par\nobreak\vspace{1pt}
\texttt{[]}\par\addvspace{5pt}
\noindent \recordkey{summary}\par\nobreak\vspace{1pt}
Implemented the frozen unreconciled\_reading\_gate arm in candidate\_source/base\_terminus.py only, preserving the existing harness byte-for-byte when it is disabled. The new runtime kwarg unreconciled\_reading\_gate defaults false and is bound in Agent.\_\_init\_\_; \_switches.armed reports the new id 'unreconciled\_reading\_gate' in addition to the three retained ids (observed\_operand\_release\_gate, require\_designated\_invocation, structured\_completion\_gate). With all four keywords false, Agent.\_run\_agent\_loop remains a single await super(); the new state is reset in Agent.\_reset\_per\_run\_state. When the new arm is armed, Agent.\_handle\_llm\_interaction also captures self.\_rsi\_gate\_raw\_reply and Agent.\_execute\_commands also accumulates self.\_rsi\_dispatched\_keystrokes. At the confirming-claim divergence (is\_task\_complete and was\_pending\_completion) the arm runs independently beside the three existing arms; the loop withholds if any arm says withhold, and the new arm's withhold re-issues the inherited confirmation message extended with one sentence naming the unreconciled reading while keeping \_pending\_completion set. The decision builds, from state already held, set (A) of (method signature, observed value token) pairs and set (B) of distinct observed value tokens for the input, and withholds once (reason unreconciled\_reading) when the claimed value's producing readings carry exactly one distinct signature (single\_signature) or when the input's readings carry more than one value and none formed differently from the claiming reading produced the claimed value (disagreeing\_readings); otherwise it releases (reading\_reconciled). Missing information (no\_claim\_value, no\_readings, no\_signature) fails open with a reported release. Every path that reaches the comparison reports exactly once through the harness-provided \_rsi\_arm\_event recorder, and nothing is reported while the switch is false or the bound is spent. The withhold is bounded once per run by \_rsi\_unreconciled\_withholds\_used capped at \_MAX\_WITHHOLDS=1; it runs no command, subprocess, or probe, so it adds no timeout. Three new pure module-level helpers (\_normalize\_method\_signature, \_extract\_value\_tokens, \_input\_identity) plus two reading-extraction helpers implement the contract's missing normalization/value-token/input-identity notions over data already held. Selftests inside candidate\_source exercise the helpers, the disabled no-op, the single-signature withhold, the reconciled release, the unknown release paths, the once-per-run bound, the switches reader, and the raw-reply capture; 'selftest ok' passes. This is an implementation report only: no conformance or performance claim is made.\par\addvspace{5pt}
\medskip\noindent\textbf{Independent-check receipt:}\par
\noindent \recordkey{status}: not\_run\par\addvspace{3pt}
\noindent \recordkey{scope}: behavior checks disabled; arms are judged by trial performance\par\addvspace{4pt}
\medskip\noindent\textbf{Full-confirmation receipt:}\par
\noindent \recordkey{status}: promotion\_ready\par\addvspace{3pt}
\noindent \recordkey{scored\_counts}\par\nobreak\vspace{1pt}
\noindent \recordkey{reference}\par\nobreak\vspace{1pt}
\noindent \recordkey{passed}: 230\enspace\textcolor{recordmuted}{\textbar}\enspace\allowbreak \recordkey{failed}: 170\enspace\textcolor{recordmuted}{\textbar}\enspace\allowbreak \recordkey{runtime\_failures}: 8\enspace\textcolor{recordmuted}{\textbar}\enspace\allowbreak \recordkey{trials}: 400\par\addvspace{3pt}\par\addvspace{5pt}
\noindent \recordkey{candidate}\par\nobreak\vspace{1pt}
\noindent \recordkey{passed}: 239\enspace\textcolor{recordmuted}{\textbar}\enspace\allowbreak \recordkey{failed}: 161\enspace\textcolor{recordmuted}{\textbar}\enspace\allowbreak \recordkey{runtime\_failures}: 7\enspace\textcolor{recordmuted}{\textbar}\enspace\allowbreak \recordkey{trials}: 400\par\addvspace{3pt}\par\addvspace{5pt}\par\addvspace{5pt}
\end{tcolorbox}

\paragraph{Reading these records.}
The pair-level feasibility judgement and the group-level proposal belong to
separate stages and can differ; both are reproduced rather than merged.
The GPT-6-sol candidate recognizes four predefined instruction patterns,
so its implementation is narrower than an arbitrary-task check generator.
The DeepSeek candidate's conflicting-readings branch has the limitations in
Appendix~\ref{app:model-behaviors}. Both examples record independent behavioral
checks as not run. DeepSeek's 239/400 confirmation is the original adoption
measurement, distinct from the later same-code repeat reported in the main
table. The research audits flag missing planned-query receipts for GPT-6-sol
and one unfetched source for DeepSeek; the displayed external-source record
has fetched-page provenance. Saved source citations do not establish the
correctness of a proposed adaptation.

\subsection{Prompt Examples}
\leavevmode\phantomsection\label{app:prompt-excerpts}%
The proposal messages for the two examples contain the same directives at
lines 3, 5, and 6, reproduced below. Their model-specific group JSON at line 4
is shown with each example above. Context inventories and absolute-path
checklists are retained in the full messages but omitted here.
The implementation prompts share an identical instruction array prefix;
that prefix and the GPT-6-sol repair instructions are reproduced in full.
These are task-level instructions, not the complete loaded skill or system prompt.

\begin{tcolorbox}[recordpanel,title={Proposal generation / saved message line 3},title after break={Proposal generation / saved message line 3 (continued)}]
\small\color{recordink}\raggedright\setlength{\baselineskip}{11.7pt}
{\color{recordmuted}\small Source: Both model-specific \path{pi/message.txt} files, line 3\par}\vspace{6pt}
Read ALL cohort members; no member cap applies. Use this accepted group wish as a hypothesis, not a conclusion:
\end{tcolorbox}

\begin{tcolorbox}[recordpanel,title={Proposal generation / saved message line 5},title after break={Proposal generation / saved message line 5 (continued)}]
\small\color{recordink}\raggedright\setlength{\baselineskip}{11.7pt}
{\color{recordmuted}\small Source: Both model-specific \path{pi/message.txt} files, line 5\par}\vspace{6pt}
Read stable-task material with suggest2\_stable. It is counterexample material only: absent recorded calls do not establish non-interference, aggregate counts do not establish a trigger, and repeat counts are not necessarily four. Do not fabricate stable quotes. Search actual precedents with the configured web tools; a network failure is not evidence of no precedents. Record wish/revise/precedent/handbook/propose through the tools, then stop. No implementation here. A proposal whose mechanism needs a signal your search plan lists under missing\_signals must produce that signal itself within allowed\_changes, for example by having the agent run its own check; otherwise record no proposal for that wish.
\end{tcolorbox}

\begin{tcolorbox}[recordpanel,title={Proposal generation / saved message line 6},title after break={Proposal generation / saved message line 6 (continued)}]
\small\color{recordink}\raggedright\setlength{\baselineskip}{11.7pt}
{\color{recordmuted}\small Source: Both model-specific \path{pi/message.txt} files, line 6\par}\vspace{6pt}
Before calling suggest2\_search\_plan (not merely before external search), use the local read tool to complete EVERY role in this checklist. Successful nonempty read results must finish before the search-plan tool call starts. For reference/dependency, read a relevant source file from any listed root, selected from that root's files inventory in the context; a directory listing is not a source read. For handbook/protocol roles, read relevant passages from an exact listed file. You need not read an entire handbook or every dependency root. One successful read can satisfy multiple roles when paths overlap. Only roles actually configured below are required; trace/cohort material is supplementary.
\end{tcolorbox}

\begin{tcolorbox}[recordpanel,title={Implementation / complete shared instructions[0:14]},title after break={Implementation / complete shared instructions[0:14] (continued)}]
\small\color{recordink}\raggedright\setlength{\baselineskip}{11.7pt}
{\color{recordmuted}\small Source: Both \path{stages/implement/.../pi_stage/prompt.txt} files, \path{/instructions/0} through \path{/instructions/13}\par}\vspace{6pt}

\begin{enumerate}[leftmargin=1.5em,itemsep=5pt,topsep=2pt]
\item You are one stage in a harness iteration pipeline. Read the frozen inputs before acting.
\item Do not edit frozen proposals, contracts, reference results/source, controller, adapters, or expectations.
\item Read each imported proposal's embedded research archives when present. Resolve borrowed against research.precedents.document; research.search\_plans.document explains the current framework and missing signals. Treat adaptation.required\_changes and validation\_needs as guidance: where a signal they ask for cannot be produced, build on the closest signal the harness has and say so in the arm description or report; that is not a reason to defer. These are research claims, not verified capabilities or permission to change frozen contracts.
\item Write the required JSON artifact to product\_path. Exit text is not an artifact or an acceptance verdict.
\item Do not install packages, access credentials, call a benchmark/model runner, or start remote jobs.
\item No performance improvement is established here. Do not claim conformance or recorder validity yourself.
\item Implement the frozen arms in candidate\_source only. Preserve the existing harness when every new arm is disabled, including retained base arms.
\item For Harbor, preserve the configured agent\_import\_path and construct each arm from its frozen runtime\_kwarg, default false. Support every retained old flag even if replaced. The configured arm\_reader\_attr must report the actual enabled switches using arm\_reader\_values (arm\_ids by default, kwargs only when explicitly configured); accepting a keyword via **kwargs without implementing it is not support. Do not modify the runtime mapping, observer, or frozen cases to fit your implementation.
\item Keep trigger predicates and actual decisions consistent. Every time an enabled arm intervenes, report it exactly once through the harness-provided recorder, and report 'unknown' when missing information stops the decision: record = getattr(self, "\_rsi\_arm\_event", None) if record is not None:     record("<arm\_id>", "triggered", reason="<short, JSON-serializable>") Do not report checks that found nothing to do, never define or replace \_rsi\_arm\_event, and never report while the arm's switch is false. These reports are the only evidence that an arm acted in a real trial.
\item Honor requires/replaces, bound intervention retries and resource use, and keep harness/infra failures distinct from model verdicts. Do not widen completion gates or invent benchmark acceptance metadata.
\item You may add selftests inside candidate\_source, but cannot edit the independently frozen cases, their adapters, or proposal scope. Use existing project tools; no actual benchmark runs.
\item Write \{status:'complete',implemented\_arms:[all frozen arm ids],deviations:[],summary:'...'\} to product\_path. If impossible, report status:'blocked' and concrete reasons. List any departure from the contract in deviations rather than hiding it; it does not end the round.
\item Only candidate\_source code may change. Your report and local notes may be written inside work\_dir.
\item No conformance case exists this round: each arm is judged by measured trial performance and by its recorder reports alone. Implement every arm's full mechanism as its contract describes, leave the harness exactly as it was while the arm is disabled, and report every intervention.
\end{enumerate}

\end{tcolorbox}

\begin{tcolorbox}[recordpanel,title={GPT-6-sol / saved implementation-repair instructions},title after break={GPT-6-sol / saved implementation-repair instructions (continued)}]
\small\color{recordink}\raggedright\setlength{\baselineskip}{11.7pt}
{\color{recordmuted}\small Source: \path{stages/implement/2/pi_stage/prompt.txt}, \path{/instructions/14} and \path{/instructions/15}\par}\vspace{6pt}

\begin{enumerate}[start=15,leftmargin=1.5em,itemsep=5pt,topsep=2pt]
\item This is implementation repair 1 of 2: the previous candidate failed a build check. Read failures\_path: check names it (source\_names: the source reads names it never defines or imports; trace\_smoke: the trace recorder could not load or record the candidate), and stderr\_tail holds the details, such as each unbound name with its lines or an import-time exception.
\item candidate\_source already holds the previous implementation; fix its code so the module imports, every name it reads is defined or imported, and every enabled arm runs. Keep the frozen contract, proposals and each arm's scope unchanged, and keep the harness exactly as before while arms are disabled.
\end{enumerate}

\end{tcolorbox}


\FloatBarrier

\section{Model-Specific Adaptation}
\label{app:model-behaviors}

This appendix describes the changes analyzed in
Section~\ref{sec:model-dependence}, separating what each mechanism does from
its measured effects. The two main models are presented first; the earlier
GPT-5.6-Terra change is included separately. Evaluation rules and numerical
results follow in Appendices~\ref{app:measurement}
and~\ref{app:supplementary-results}.

\subsection{GPT-6-sol}
\label{app:gpt6-behaviors}

The three retained changes improve execution continuity and the evidence
returned to the model. Their behavioral locations follow the seed Handbook:
session liveness is checked by the Interaction main loop; persisted changes
are presented during Observation construction and output truncation; and
completion-specific context is supplied during the Completion confirmation
handshake.

\paragraph{\normalfont\bfseries Completion-claim context.}
When the agent declares completion, the confirmation message includes evidence
of the work performed earlier in the run, rather than only an idle terminal
prompt. This modification targets the Completion confirmation handshake,
supported by observation construction and the recorded execution history.
Its implementation setting is \code{completion_claim_context}.

\paragraph{\normalfont\bfseries Persisted workspace feedback.}
The harness presents changes actually written to files as a workspace diff.
The feedback is grounded in persisted file contents rather than terminal echo,
giving the agent a view of what changed. Its primary behavioral location is
Observation construction and output truncation, with file inspection supplying
the underlying evidence. Its setting is \code{persisted_workspace_diff}.

\paragraph{\normalfont\bfseries Session-loss recovery.}
When the terminal session disappears, the harness recovers the session and
informs the agent of the resulting state loss. This changes the Interaction
main loop's handling of session liveness, with tmux session operations and
observation construction supporting recovery and its feedback. Its setting is
\code{single_session_loss_recovery}.

\subsection{DeepSeek-V4-Flash}
\label{app:deepseek-behaviors}

In the seed harness, the first completion declaration requests confirmation,
and a second consecutive declaration permits termination. The four adopted
modifications add checks to this handshake. In Figures~\ref{fig:outcome-distribution}
and~\ref{fig:deepseek-evolution}, SE denotes structured evidence, DI designated
invocation, OO observed operand, and IR independent reading. The corresponding
behaviors are detailed below. Each check can withhold completion
at most once per run, providing a bounded opportunity for correction rather
than enforcing an indefinitely repeated verification requirement.

\paragraph{\normalfont\bfseries Primary and supporting stages.}
The primary stage is where a modification changes a harness decision or action;
supporting stages acquire its inputs or maintain the required state. For
example, designated-invocation verification reads command history recorded
during dispatch but changes the completion decision during the handshake.
All four modifications therefore target Completion confirmation handshake,
with their supporting stages listed in Table~\ref{tab:deepseek-behaviors}.

\begin{table}[!htbp]
\centering\small
\renewcommand{\arraystretch}{1.15}
\setlength{\tabcolsep}{3pt}
\begin{tabularx}{\linewidth}{@{}>{\raggedright\arraybackslash}p{0.30\linewidth}
  >{\raggedright\arraybackslash}p{0.24\linewidth}Y@{}}
\toprule
Behavior & Primary stage & Supporting stages \\
\midrule
Executable completion verification (SE) & Completion confirmation handshake &
LLM query and context rescue; Response parsing and command extraction:
preserve and parse the verification command and expected exit code. \\
Designated-invocation verification (DI) & Completion confirmation handshake &
Command dispatch and timeout handling: record dispatched commands. \\
Completion-outcome observation (OO) & Completion confirmation handshake &
LLM query and context rescue; Response parsing and command extraction:
obtain verification fields for the environment probe. \\
Independent-reading corroboration (IR) & Completion confirmation handshake &
Command dispatch and timeout handling; Observation construction and output
truncation: record commands and construct observations containing reading
results. \\
\bottomrule
\end{tabularx}
\caption{Behavioral modifications for DeepSeek-V4-Flash and their Handbook mappings.}
\label{tab:deepseek-behaviors}
\end{table}

\paragraph{\normalfont\bfseries Executable completion verification (SE).}
The model supplies a verification command and its expected exit code in the
completion declaration. The harness executes the command and compares its
exit code with the expected value. A match permits
this check to release completion; missing fields, execution failure, or a
mismatch produce one request to continue working. This makes the model's
completion evidence executable by the harness, while preserving the bounded
withholding rule.

\paragraph{\normalfont\bfseries Designated-invocation verification (DI).}
When a task explicitly specifies an invocation, the harness checks whether it
appears in the dispatched-command history. It conservatively extracts the
interpreter, script, and arguments from the task instruction, and accommodates
absolute paths and relative script paths after a directory change through
basename matching. The check triggers only when an alternative invocation has
been observed but the designated invocation has not. It then requests that the
model execute the specified invocation before confirming completion again.

\paragraph{\normalfont\bfseries Completion-outcome observation (OO).}
On the second completion declaration, the harness executes the supplied
verification command through the session's environment interface, with a
30-second probe limit. The check compares an observed integer exit code with
the expected value and records a successful match as well as a withholding
decision. Missing evidence, a non-integer result, or a mismatch can withhold
completion once. If the sandbox call raises an exception, this check fails
open, so the probe exception does not itself turn into a task failure.

\paragraph{\normalfont\bfseries Independent-reading corroboration (IR).}
This check uses existing execution history rather than launching another
command. It relates dispatched commands, normalized reading-method signatures,
input objects, and values extracted from subsequent observations. If a reading
used to support completion is backed by only one method, the harness requests
a different form of inspection or a second independent reading method.
Repeating the same method is not treated as independent corroboration.

\paragraph{\normalfont\bfseries Implementation boundaries.}
Executable completion verification and completion-outcome observation are
separate checks and can execute the same verification command twice. Their
effects are therefore overlapping. Independent-reading corroboration primarily
detects reliance on a single reading method: the current evidence-set
construction makes the separate conflicting-readings branch largely
unreachable. We consequently do not describe it as a general consistency
checker. These are implementation boundaries, not estimates of the
individual mechanisms' causal contributions. The full cumulative measurements
and their evaluation provenance are reported in
Appendix~\ref{app:cumulative-results}.

\subsection{GPT-5.6-Terra}
\label{app:gpt56-behavior}

The earlier GPT-5.6-Terra modification targets Command dispatch and timeout
handling. A command can contain a literal tab as input data; if tmux interprets
it as a special key event, the terminal receives different input from the
model's intended command. Literal control-character transport preserves the
character as data during dispatch. This changes command transport rather
than completion confirmation. Its results are reported separately in
Appendix~\ref{app:gpt56-results} and do not establish the changes needed by
GPT-6-sol.


\FloatBarrier

\section{Evaluation Methodology}
\label{app:measurement}

This appendix supplements Section~\ref{sec:experimental-setup} with scoring,
baseline, and version-reporting rules. Numerical results follow in
Appendix~\ref{app:supplementary-results}.

\subsection{Repeated Evaluation}
\leavevmode\phantomsection\label{app:scoring}%
The default $r=4$ gives 400 trial slots for TBH and 356 for TB2.1.
The metrics in Section~\ref{sec:metrics} require a complete grid. Passes score
$1$; task failures and completed runtime failures with valid records score $0$,
retaining their harness, infrastructure, or unknown-error classification.
Missing, cancelled, or identity-mismatched measurements are $\bot$ and cannot
be dropped to make an incomplete grid valid. Blank table entries mean
unreported results, not failures.

Aggregate outcome proportions do not identify which initially stable tasks
regressed; paired per-task outcomes are needed.

\subsection{Comparison Protocol}
\leavevmode\phantomsection\label{app:baseline-policy}%
Before evaluating a candidate, the policy specifies the comparison baseline.
For stage $s$ on task set $S$, it selects either historical reference runs or
a newly measured incumbent control:
\begin{equation}
\widehat p_B^{\,s}(S)=
\begin{cases}
\widehat p(R_t|_S), & b_s=\mathrm{historical},\\
\widehat p(H_t;S,r_s), & b_s=\mathrm{fresh}.
\end{cases}
\label{eq:baseline}
\end{equation}
Here $r_s$ is the stage's repetition count. Historical comparisons use their
stored repetitions and corresponding denominator. The controller defaults
to historical reuse; fresh controls are optional. Historical source identity
does not establish unchanged service or environment conditions, so baseline
provenance remains part of the evaluation record.

Adoption requires a valid, complete full evaluation with aggregate pass rate
above the declared baseline. Previously stable tasks may regress despite an
overall gain, so reliability is reported separately from the adoption score.

\subsection{Result Reporting}
\label{app:version-reporting}

The main comparison, cumulative curves, and transfer experiment refer to
explicitly identified harness versions. On DeepSeek-V4-Flash, TBH uses the
final four-mechanism version (+ IR). Table~\ref{tab:main-results} and
Figure~\ref{fig:transfer-comparisons} report the best observed TB2.1
three-mechanism version (+ OO), selected by its TB2.1 score. This is distinct
from directly transferring the final version selected on TBH. The full
sequence in Appendix~\ref{app:cumulative-results} keeps both versions visible.
TB2.1 does not inform harness modifications; selection by its observed score
is nevertheless part of this reporting protocol.

The four non-base TBH versions use the original $100\times4$ grid.
The seed's TBH result is 227/400 passes (56.75\%), with 37 all-success,
34 flipping, and 29 all-failure tasks. For the first three
non-base TB2.1 versions, controller-timeout slots were rerun and merged with
the other results; the final version uses a complete rerun. These differences
in evaluation assembly are retained when interpreting the cumulative sequence.
The curves measure combined harness versions rather than isolated causal
effects of individual mechanisms.


\FloatBarrier

\section{Supplementary Results}
\label{app:supplementary-results}

The following results extend the main experiments in order: main-comparison
trial counts, cumulative DeepSeek-V4-Flash versions, component ablations,
and cross-model transfer.
The earlier GPT-5.6-Terra experiment follows separately. Mechanisms and
measurement rules are defined in Appendices~\ref{app:model-behaviors}
and~\ref{app:measurement}.

\subsection{GPT-6-sol}
\label{app:gpt6-results}

Table~\ref{tab:gpt6-execution} gives the GPT-6-sol pass counts underlying
Table~\ref{tab:main-results}, retaining the full $4N$ denominator. Its
ablations remain unreported. Cross-model
transfer measurements appear in Appendix~\ref{app:cross-model-results}.

\begin{table}[!htbp]
\centering\small
\setlength{\tabcolsep}{10pt}
\begin{tabular}{@{}lrr@{}}
\toprule
Harness & TBH & TB2.1 \\
\midrule
Terminus-2 & 201 / 400 & 280 / 356 \\
ModularRSI & 203 / 400 & 280 / 356 \\
AHE Harness & 219 / 400 & 278 / 356 \\
\method{} & 222 / 400 & 291 / 356 \\
\bottomrule
\end{tabular}
\caption{GPT-6-sol passed / total trials.}
\label{tab:gpt6-execution}
\end{table}

\subsection{Evolution Results}
\leavevmode\phantomsection\label{app:cumulative-results}%
Table~\ref{tab:cumulative-results} gives the values plotted in
Figures~\ref{fig:outcome-distribution} and~\ref{fig:deepseek-evolution}.
Each ``+'' adds one of the mechanisms defined in
Appendix~\ref{app:model-behaviors}: structured evidence (SE), designated
invocation (DI), observed operand (OO), and independent reading (IR).
These are cumulative versions, not controller round identifiers.

\begin{table}[!htbp]
\centering\small
\setlength{\tabcolsep}{7pt}
\begin{tabular}{@{}lrrrrr@{}}
\toprule
Version & Pass@1 & Pass$^4$ & Pass@4 & Flip & All-failure \\
\midrule
\multicolumn{6}{@{}l}{\textbf{TBH}} \\
Base & 56.75 & 37.00 & 71.00 & 34.00 & 29.00 \\
+ SE & 57.75 & 39.00 & 70.00 & 31.00 & 30.00 \\
+ DI & 56.75 & 42.00 & 71.00 & 29.00 & 29.00 \\
+ OO & 57.50 & 41.00 & 70.00 & 29.00 & 30.00 \\
+ IR & 64.00 & 51.00 & 73.00 & 22.00 & 27.00 \\
\midrule
\multicolumn{6}{@{}l}{\textbf{TB2.1}} \\
Base & 72.19 & 58.43 & 86.52 & 28.09 & 13.48 \\
+ SE & 77.81 & 65.17 & 88.76 & 23.60 & 11.24 \\
+ DI & 77.81 & 66.29 & 88.76 & 22.47 & 11.24 \\
+ OO & 79.49 & 67.42 & 89.89 & 22.47 & 10.11 \\
+ IR & 75.28 & 65.17 & 85.39 & 20.22 & 14.61 \\
\bottomrule
\end{tabular}
\caption{Cumulative DeepSeek-V4-Flash results ($r=4$; \%), matching
Figures~\ref{fig:outcome-distribution} and~\ref{fig:deepseek-evolution}.}
\label{tab:cumulative-results}
\end{table}

The final + IR version achieves the highest TBH Pass@1 and Pass$^4$.
On TB2.1, the + OO version has the highest observed values, whereas direct
transfer of the TBH-selected + IR version improves Pass@1 and Pass$^4$ over
the seed by 3.09\% and 6.74\%, respectively. The version-selection
and rerun procedures are specified in Appendix~\ref{app:version-reporting}.
The category proportions show aggregate distributions, not transitions of
particular tasks.

\subsection{Ablation Results}
\label{app:ablation-results}

Table~\ref{tab:ablation-results} complements the TBH comparisons in
Figure~\ref{fig:ablation-comparisons}
and the comparisons in Section~\ref{sec:ablations}. HE denotes harness traces
and the Handbook; ER denotes external design research. Both ablations retain
the remaining components. The full method uses + IR on TBH and + OO on TB2.1,
following the version-reporting protocol above.

\begin{table}[!htbp]
\centering\small
\setlength{\tabcolsep}{6pt}
\begin{tabular}{@{}llrrrr@{}}
\toprule
Benchmark & Method & Pass@1 & Pass$^4$ & Pass@4 & Flip \\
\midrule
TBH & \method{} & 64.00 & 51.00 & 73.00 & 22.00 \\
 & \method{} w/o HE & 57.00 & 43.00 & 72.00 & 29.00 \\
 & \method{} w/o ER & 58.50 & 45.00 & 68.00 & 23.00 \\
\midrule
TB2.1 & \method{} & 79.49 & 67.42 & 89.89 & 22.47 \\
 & \method{} w/o HE & 76.40 & 62.92 & 88.76 & 25.84 \\
 & \method{} w/o ER & 76.69 & 64.04 & 88.76 & 24.72 \\
\bottomrule
\end{tabular}
\caption{DeepSeek-V4-Flash ablations ($r=4$; \%). These values support
Section~\ref{sec:ablations} and are kept separate from the main comparison table.}
\label{tab:ablation-results}
\end{table}

\subsection{Cross-Model Transfer}
\leavevmode\phantomsection\label{app:cross-model-results}%
We apply each model's final evolved harness to the other model, keeping the
transferred harness fixed, and evaluate each task four times on both
benchmarks. Table~\ref{tab:cross-model-transfer} reports these cross-model
evaluations. This differs from cross-benchmark generalization, which keeps
the execution model unchanged.

\begin{table}[!htbp]
\centering\small
\setlength{\tabcolsep}{4pt}
\begin{tabular}{@{}llrrrr@{}}
\toprule
Benchmark & Passed / total & Pass@1 & Pass$^4$ & Pass@4 & Flip \\
\midrule
\multicolumn{6}{@{}l}{\textbf{DeepSeek-V4-Flash harness $\rightarrow$ GPT-6-sol}} \\
TBH & 211 / 400 & 52.75 & 42.00 & 62.00 & 20.00 \\
TB2.1 & 270 / 356 & 75.84 & 60.67 & 89.89 & 29.21 \\
\midrule
\multicolumn{6}{@{}l}{\textbf{GPT-6-sol harness $\rightarrow$ DeepSeek-V4-Flash}} \\
TBH & 233 / 400 & 58.25 & 45.00 & 69.00 & 24.00 \\
TB2.1 & 260 / 356 & 73.03 & 62.92 & 83.15 & 20.22 \\
\bottomrule
\end{tabular}
\caption{Cross-model transfer of final evolved harnesses ($r=4$).
Pass metrics and Flip are in \%; task counts use 100 tasks on TBH
and 89 on TB2.1. Arrows indicate harness source model $\rightarrow$
execution model.}
\label{tab:cross-model-transfer}
\end{table}

Figure~\ref{fig:cross-model-transfer} uses the final + IR harness for
DeepSeek-V4-Flash on both benchmarks, rather than the best-observed + OO
version in Table~\ref{tab:main-results}. GPT-6-sol's own-adapted TB2.1
reference is the 291/356 \method{} result in Table~\ref{tab:gpt6-execution}.

\subsection{GPT-5.6-Terra}
\leavevmode\phantomsection\label{app:gpt56-results}%
These earlier experiments use GPT-5.6-Terra, a different model from GPT-6-sol
in the main evaluation. Its results are kept separate from the GPT-6-sol rows in
Table~\ref{tab:main-results}. The command-transport modification is described
in Appendix~\ref{app:gpt56-behavior}.

\begin{table}[!htbp]
\centering\small
\setlength{\tabcolsep}{1pt}
\renewcommand{\arraystretch}{1.15}
\begin{tabularx}{\linewidth}{@{}>{\raggedright\arraybackslash}p{0.21\linewidth}
  *{4}{>{\centering\arraybackslash}X}@{\hspace{6pt}}
  *{4}{>{\centering\arraybackslash}X}@{}}
\toprule
\multirow{2}{*}{Harness} & \multicolumn{4}{c}{\textbf{TBH}}
& \multicolumn{4}{c}{\textbf{TB2.1}} \\
\cmidrule(lr){2-5}\cmidrule(lr){6-9}
& Pass@1 & Pass$^4$ & Pass@4 & Flip$\downarrow$
& Pass@1 & Pass$^4$ & Pass@4 & Flip$\downarrow$ \\
\midrule
Terminus-2 & 41.25 & 22.00 & 58.00 & 37.00 & 58.10 & 34.83 & 75.30 & 44.94 \\
\rowcolor{metabg!8}
\method{} & \resultgap{\textbf{43.50}}{+2.25} & \resultgap{\textbf{29.00}}{+7} & \resultgap{\textbf{59.00}}{+1} & \resultgap{\textbf{30.00}}{-7} & \resultgap{\textbf{62.92}}{+4.82} & \resultgap{\textbf{43.82}}{+8.99} & \resultgap{\textbf{80.90}}{+5.6} & \resultgap{\textbf{37.08}}{-7.86} \\

\bottomrule
\end{tabularx}
\caption{Historical GPT-5.6-Terra results ($r=4$; \%), separate from
GPT-6-sol. Coral annotations show absolute changes from Terminus-2.}
\label{tab:gpt56-results}
\end{table}



\FloatBarrier

\section{Scope and Limitations}

\paragraph{Execution evidence and analysis coverage.}
\leavevmode\phantomsection\label{sec:limitations}%
Successful runs provide references for failure analysis, but differences may
also reflect sampling, environment variation, or task difficulty. Discovery
from traceable flipping tasks excludes consistently failing tasks and failures
before recording. The monitor covers selected boundaries, and the Handbook
may omit behavior or become stale. Neither source establishes causality.

\paragraph{Candidate changes and regression risks.}
External designs may require unavailable interfaces or signals. Behavioral
checks cover finite cases, while combined-candidate evaluation does not
isolate individual mechanisms. Aggregate-score adoption can retain changes
that harm previously stable tasks; aggregate outcome proportions cannot
quantify these regressions without per-task comparisons.

\paragraph{Evaluation and transfer.}
Repeated development-set selection can exploit noise, and historical
baselines are sensitive to service or environment drift. Retained-score
trends therefore require independent evaluation. The reporting protocol in
Appendix~\ref{app:version-reporting} distinguishes target-selected versions
from frozen final-version transfer. Cross-model swapping shows model
dependence for the tested harnesses, without ruling out transfer of individual
mechanisms between other models.

\paragraph{Scope and resource costs.}
Other harnesses require compatible monitoring, behavioral mappings, and
evaluation adapters. Long runs, external research, and repeated evaluation
incur inference and storage costs; configured concurrency is not measured
speedup. LLMCarbon motivates broader resource accounting
\citep{faiz2024llmcarbon}, but emissions are not measured here.
The loop keeps model weights fixed and does not learn its outer controller.
Training-side methods, including work on asynchronous policy mismatch
\citep{yang2026stalestable}, are complementary directions for future study.



\end{document}